\documentclass[11pt,twoside]{article}
\usepackage{fancyhdr}
\usepackage[bookmarks=false]{hyperref}
\usepackage{amsfonts,epsfig,graphicx}
\usepackage{afterpage}
\usepackage{amsmath,amssymb,amsthm} 
\usepackage{fullpage}
\usepackage[T1]{fontenc} 
\usepackage{epsf} 
\usepackage{graphics} 
\usepackage{amsfonts,amsmath}
\usepackage{natbib} 
\usepackage{psfrag,xspace}
\usepackage{color,etoolbox}
\usepackage[usenames,dvipsnames]{xcolor}

\hypersetup{colorlinks,
citecolor={RoyalBlue},
urlcolor={RoyalBlue},
linkcolor={RoyalBlue}}

\theoremstyle{plain}
\newtheorem{theorem}{Theorem}[section]

\newtheorem{proposition}[theorem]{Proposition}

\theoremstyle{definition}

\newtheorem{assumption}[theorem]{Assumption}
\newtheorem{example}[theorem]{Example}

\usepackage{times}
\usepackage{subcaption}
\usepackage{makecell}
\usepackage{multirow}
\usepackage{outlines}
\usepackage{cancel}
\usepackage{footnote}
\usepackage{pifont}
\usepackage{rotating}
\usepackage{enumitem}
\usepackage{booktabs}  
\usepackage{tikz}
\usetikzlibrary{bayesnet}

\newcommand{\inparen}[1]{\left(#1\right)}
\newcommand{\incurly}[1]{\left\{#1\right\}}
\newcommand{\insquare}[1]{\left[#1\right]}

\newcommand{\norm}[1]{\left\lVert#1\right\rVert}

\newcommand{\indicator}[1]{\mathbf{1}\inparen{#1}}

\DeclareMathOperator*{\argmax}{argmax}

\newcommand{\R}{\mathbb{R}}

\newcommand{\calD}{\mathcal{D}}

\newcommand{\calN}{\mathcal{N}}

\newcommand{\calX}{\mathcal{X}}
\newcommand{\calY}{\mathcal{Y}}

\newcommand{\Esymb}{\mathbb{E}}
\newcommand{\Psymb}{\mathbb{P}}
\newcommand{\Vsymb}{\mathsf{Var}}

\newcommand{\prob}[1]{\Psymb\left({#1}\right)}

\newcommand{\ex}[1]{\Esymb\left[{#1}\right]}
\newcommand{\Ex}[2]{\Esymb_{{#1}}\left[{#2}\right]}

\newcommand{\Var}[2]{\Vsymb_{{#1}}\left[{#2}\right]}

\newcommand{\condprob}[2]{\Psymb\left({#1}\;\middle\vert\;{#2}\right)}
\newcommand{\condex}[2]{\Esymb\left[{#1}\;\middle\vert\;{#2}\right]}

\newcommand{\indep}{\perp \!\!\! \perp}

\newcommand{\sfA}{\mathsf{A}}
\newcommand{\sfB}{\mathsf{B}}

\newcommand{\sfr}{\mathsf{r}}
\newcommand{\sfs}{\mathsf{s}}

\newcommand{\plugin}{\mathsf{pi}}
\newcommand{\ipw}{\mathsf{ipw}}
\newcommand{\dr}{\mathsf{dr}}
\newcommand{\influence}{\mathsf{IF}}

\newcommand{\email}[1]{\href{mailto:#1}{\textcolor{black}{\texttt{#1}}}}

\newcommand{\newlyadded}[1]{{#1}}

\usepackage{etoolbox}
\newtoggle{compact}
\togglefalse{compact}

\let\svthefootnote\thefootnote
\newcommand\freefootnote[1]{%
  \let\thefootnote\relax%
  \footnotetext{#1}%
  \let\thefootnote\svthefootnote%
}

\begin{document}


\title{%
\vspace{-2em}
{\bf Counterfactually Comparing Abstaining Classifiers}\freefootnote{Accepted to the \emph{37th Conference on Neural Information Processing Systems (NeurIPS 2023)}.}
\vspace{1em}
}
\author{%
    {\bf Yo Joong Choe}\thanks{This work was submitted while this author was at Carnegie Mellon University.} \\
    Data Science Institute \\
    University of Chicago \\
    \email{yjchoe@uchicago.edu}
    \and
    {\bf Aditya Gangrade} \\
    Department of EECS \\
    University of Michigan \\
    \email{aditg@umich.edu} \\
    \and
    {\bf Aaditya Ramdas} \\
    Department of Statistics and Data Science \\
    Machine Learning Department \\
    Carnegie Mellon University \\
    \email{aramdas@cmu.edu} 
    \vspace{1em}
}
\date{\normalsize\today}

\maketitle

\vspace{-1em}
\begin{abstract}
    \normalsize 
    Abstaining classifiers have the option to abstain from making predictions on inputs that they are unsure about.
These classifiers are becoming increasingly popular in high-stakes decision-making problems, as they can withhold uncertain predictions to improve their reliability and safety.
When \emph{evaluating} black-box abstaining classifier(s), however, we lack a principled approach that accounts for what the classifier would have predicted on its abstentions.
\newlyadded{These missing predictions matter when they can eventually be utilized, either directly or as a backup option in a failure mode.}
In this paper, we introduce a novel approach and perspective to the problem of evaluating and comparing abstaining classifiers by treating abstentions as \emph{missing data}.
Our evaluation approach is centered around defining the \emph{counterfactual score} of an abstaining classifier, defined as the expected performance of the classifier had it not been allowed to abstain.
We specify the conditions under which the counterfactual score is identifiable: if the abstentions are stochastic, and if the evaluation data is independent of the training data (ensuring that the predictions are \emph{missing at random}), then the score is identifiable.
Note that, if abstentions are deterministic, then the score is unidentifiable because the classifier can perform arbitrarily poorly on its abstentions.
Leveraging tools from observational causal inference, we then develop nonparametric and doubly robust methods to efficiently estimate this quantity under identification. 
Our approach is examined in both simulated and real data experiments.

\end{abstract}

\clearpage
\tableofcontents
\clearpage


\section{Introduction}\label{sec:introduction}

Abstaining classifiers \citep{chow1957optimum,elyaniv2010foundations}, also known as selective classifiers or classifiers with a reject option, are classifiers that have the option to abstain from making predictions on certain inputs.
As their use continues to grow in safety-critical applications, such as medical imaging and autonomous driving, it is natural to ask how a practitioner should \emph{evaluate and compare} the predictive performance of abstaining classifiers under black-box access to their decisions. 

In this paper, we introduce the \emph{counterfactual score} as a new evaluation metric for black-box abstaining classifiers.
The counterfactual score is defined as the expected score of an abstaining classifier's predictions, \emph{had it not been allowed to abstain}.
This score is of intrinsic importance when the potential predictions on abstaining inputs are relevant.
We proceed with an illustrative example:
\begin{example}[Free-trial ML APIs]\label{ex:api}
Suppose we compare different image classification APIs. 
Each API has two versions: a free version that abstains, and a paid one that does not.
Before paying for the full service, the user can query the free version for up to $n$ predictions on a user-provided dataset, although it may choose to reject any input that it deems as requiring the paid service.
Given two such APIs, how can the practitioner determine which of the two paid (non-abstaining) versions would be better on the population data source, given their abstaining predictions on a sample?
\end{example}
Example~\ref{ex:api} exhibits why someone using a black-box abstaining classifier would be interested in its counterfactual score: the user may want to reward classifiers whose hidden predictions are also (reasonably) good, as those predictions may be utilized in the future.
This is a hypothetical example, but we can imagine other applications of abstaining classifiers where the counterfactual score is meaningful.
In Appendix~\ref{sec:examples}, we give three additional examples, including safety-critical applications where the hidden predictions may be used as a backup option in a failure mode.

To formally define, identify, and estimate the counterfactual score, we cast the evaluation problem in \citet{rubin1976inference}'s missing data framework and treat abstentions as \emph{missing predictions}.
This novel viewpoint directly yields nonparametric methods for estimating the counterfactual score of an abstaining classifier, drawing upon methods for causal inference in observational studies~\citep{rubin1974estimating,robins1994estimation,vandervaart2000asymptotic}, and represents an interesting yet previously unutilized theoretical connection between selective classification, model evaluation, and causal inference. 

The identification of the counterfactual score is guaranteed under two standard assumptions: the missing at random (MAR) condition, which is satisfied as long as the evaluation data is independent of the classifier (or its training data), and the positivity condition.
As with standard missing data settings, both the MAR and positivity conditions are necessary for identification.
We later discuss each condition in detail, including when the positivity condition is met and how a policy-level approach may be necessary for safety-critical applications.

\iftoggle{compact} {
    \begin{wrapfigure}{r}{0.47\textwidth}
        \vspace{-1.5em}
        \begin{center}
            \includegraphics[width=0.47\textwidth]{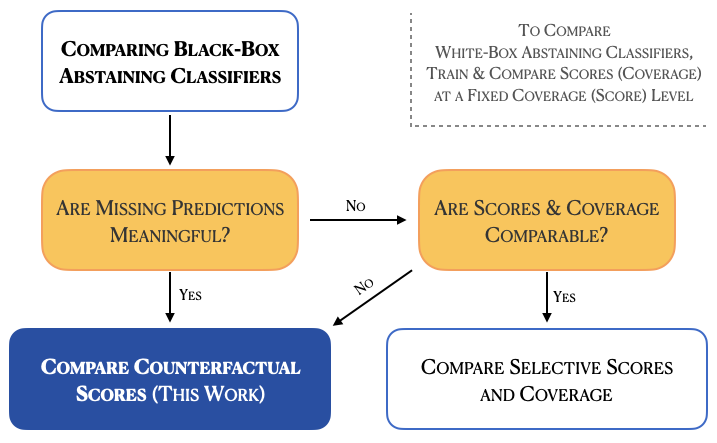}
        \end{center}
        \caption{A schematic flowchart of comparing abstaining classifiers.
        In a black-box setting where the evaluator does not have access to the training algorithms or the resources to train them, the task can be viewed as a nontrivial missing data problem. 
        This work proposes the counterfactual score as an evaluation metric.}
        \label{fig:figure1}
    \end{wrapfigure}
}{
    \begin{figure}[t]
        \begin{center}
            \includegraphics[width=0.62\textwidth]{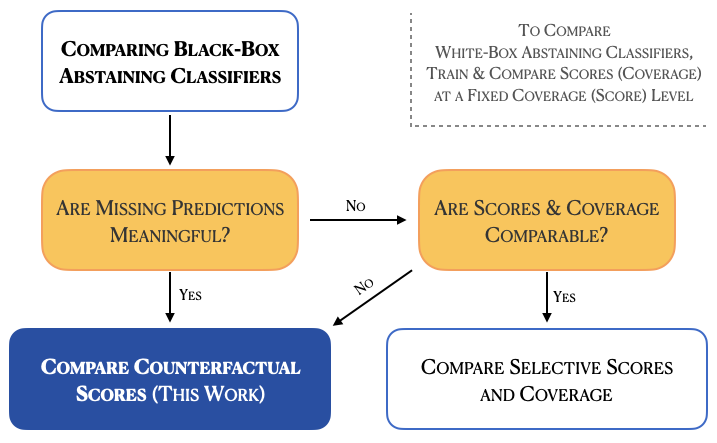}
        \end{center}
        \vspace{-\baselineskip}
        \caption{A schematic flowchart for comparing abstaining classifiers.
        In a black-box scenario where the evaluator does not have access to the underlying abstention mechanism or the base classifier, the task can be viewed as a nontrivial missing data problem. 
        This work proposes the counterfactual score, which accounts for the missing predictions, as an evaluation metric.}
        \label{fig:figure1}
    \end{figure}
}

The counterfactual score can be viewed as an alternative to the selective score (mean score on nonabstentions) and the coverage (1 minus the abstention rate)~\citep{elyaniv2010foundations}, which are the main existing metrics for evaluating black-box abstaining classifiers. 
As a two-dimensional metric, comparison on the basis of these is non-trivial. 
A common approach is to assume a fixed cost for each abstention~\citep{chow1970optimum}, but this is not always satisfactory since determining how to weigh abstentions and errors against one another is a nontrivial question. 
Thus, in settings such as Example~\ref{ex:api}, the notion of {counterfactual score} becomes necessary.
Importantly, selective scores are not representative of the counterfactual performance, except in the (unrealistic) case wherein predictions are missing completely at random (MCAR).\footnote{MCAR means the missing observations are simply a uniformly random subset of all observations, independently of the input/output. In contrast, MAR means there can be systematic differences between the missing and observed values, but these can be explained by the input. Our method only requires MAR.} 
Figure~\ref{fig:figure1} gives an overview of scenarios where different metrics may be appropriate to compare abstaining classifiers.

The counterfactual score also offers practical benefits when comparing abstaining classifiers. 
Counterfactual scores are comparable even if abstaining classifiers are tuned to different abstention rates or selective scores.
Moreover, compared to evaluation methods using the selective score-coverage curve (equivalent to re-training the classifier several times at different score/coverage levels), estimating the counterfactual score does not require re-training the classifier. 
Instead, we only need to estimate a pair of nuisance functions that can be learned using the observed predictions (nonabstentions) in the evaluation set. 
Let us further note that the setup is applicable generally to any form of prediction that can be scored, including regression and structured prediction.
In this paper, we restrict our attention to classification for concreteness, as it is the most well-studied abstention framework.

\paragraph{Summary of contributions} We first formalize the problem of comparing abstaining classifiers as a missing data problem and introduce the counterfactual score as a novel metric for abstaining classifiers. 
Next, we discuss how the counterfactual score can be identified under the MAR and positivity conditions.
Then, we develop efficient nonparametric estimators for the counterfactual scores and their differences, namely doubly robust confidence intervals (DR CI). 
Finally, we validate our approach in simulated and real-data experiments. 


\paragraph{Related work on abstaining classifiers} 
The \emph{training} of abstaining classifiers has seen significant interest in the literature~\citep[e.g.,][]{chow1970optimum,pietraszek2005optimizing,bartlett2008classification,cortes2016boosting,geifman2017selective,geifman2019selectivenet,gangrade2021selective}.
We refer to~\citet{hendrickx2021machine,zhang2023survey} for recent surveys.
For \emph{evaluation}, aside from using some combination of selective score and coverage, the most pertinent reference is the work of \citet{condessa2017performance}, who propose the metric of `classifier quality' that is a somewhat inverse version of our counterfactual accuracy. 
This metric is the sum of the prediction accuracy when the classifier predicts, and prediction \emph{inaccuracy} when it abstains, the idea being that if a classifier is not abstaining needlessly, then it must hold that the underlying predictions on points it abstains on are very poor. 
While this view is relevant to the training of abstention rules, it is at odds with black-box settings where the underlying predictions may still be executed even when the method abstains, motivating the counterfactual score. 
In terms of estimating the score, \citet{condessa2017performance} do not discuss the black-box setting that requires estimating a counterfactual, as they assume full knowledge of the classifier's predictions.

\paragraph{Related work on missing data, causal inference, and doubly robust estimation} 
Our main approach to the estimability of counterfactual scores is driven by a reduction to an inference problem under missing data (or censoring)~\citep{rubin1976inference,little2019statistical}.
A missing data problem can be viewed equivalently as a causal inference problem in an observational study~\citep{rubin1976inference,pearl2000models,shpitser2015missing,ding2018causal}, and there exist well-established theories and methods for both identifying the counterfactual quantity of interest and efficiently estimating the identified target functional.
For \emph{identification}, unlike in some observational settings for estimating treatment effects, our setup straightforwardly satisfies the standard assumption of consistency (non-interference).
The MAR assumption is satisfied as long as independent evaluation data is used, and the positivity assumption translates to abstentions being stochastic.
We discuss various implications of these conditions in Sections~\ref{sec:identification_cfscore} and~\ref{sec:discussion}.
For \emph{estimation}, efficient methods for estimating targets such as the average treatment effect (ATE) have long been studied under semiparametric and nonparametric settings.
Doubly robust (DR) estimators, in particular, are known to achieve the asymptotic minimax lower bound on the mean squared error.
For details, we refer the reader to~\citet{bickel1993efficient,robins1994estimation,vandervaart2000asymptotic,vandervaart2002semiparametric,vanderlaan2003unified,bang2005doubly,tsiatis2006semiparametric,chernozhukov2018double,kennedy2022semiparametric}. 
Unlike the standard ATE estimation setup in observational studies, our setup contrasts two separate causal estimands, the counterfactual scores of the two competing classifiers, that operate under their distinct missingness mechanisms.


\section{Definition and identification of the counterfactual score}\label{sec:cfscore}

We formulate the problem of evaluating and comparing abstaining classifiers under the missing data framework~\citep{rubin1976inference}. 
We follow the standard approach of defining the target parameter (Section~\ref{sec:target_cfscore}), identifying it with observable quantities (Section~\ref{sec:identification_cfscore}), and estimating the identified parameter using data (Section~\ref{sec:estimation_cfscore}). 
In each step, we first consider evaluating one abstaining classifier and then extend to comparing two abstaining classifiers.
Below, $\calX$ denotes the input space and $\calY = \{1, \dotsc, C\}$ is the set of possible classes, while $\Delta^{C-1}$ denotes the $C$-dimensional probability simplex on $\calY$. 

\paragraph{Abstaining classifiers} We define an abstaining classifier as a pair of functions $(f,\pi)$, representing its \emph{base classifier} $f:\calX \to \Delta^{C-1}$ and \emph{abstention mechanism} $\pi:\calX \to [0,1]$, respectively. 
Given a query $X$, the classifier first forms a preliminary (probabilistic) prediction $f(X)$. 
Then, potentially using the output $f(X)$, the classifier determines $\pi(X)$, i.e., the abstention probability. 
Using $\pi(X)$, the classifier then makes the binary abstention decision $R \mid \pi(X) \sim \mathsf{Ber}(\pi(X))$, so that if $R=1$ (``rejection''), the classifier abstains on the query, and if $R = 0$, it reveals its prediction $f(X)$. 
In some cases, we will explicitly define the source of randomness $\xi$ (independent of the data) in deciding $R$, such that $R = \sfr(\pi(X), \xi)$ for a deterministic function $\sfr$.\footnote{Specifically, let $\xi \sim \mathsf{Unif}[0,1]$ and $R = \indicator{\xi \leq \pi(X)}$. Then, $R$ is a function of only $\pi(X)$ and $\xi$.}
Neither $f$ nor $\pi$ is assumed to be known to the evaluator, modeling the black-box access typically available to practitioners.

\paragraph{Scoring rules (Higher scores are better.)} 
We measure the quality of a prediction $f(x)$ for a label $y$ via a positively oriented \emph{scoring rule} $\sfs: \Delta^{C-1}\times \calY \to \R$. 
One simple scoring rule is classification accuracy, i.e., $\sfs(f(x),y) = \indicator{\argmax_{c \in \calY} f(x)_c = y}$, but a plethora of scores exist in the literature, such as the \citet{brier1950verification} score: $\sfs(f(x),y) = 1 - \sum_{c \in \calY} (f(x)_c - \indicator{y = c})^2.$ 

\paragraph{The evaluation setup} 
For each labeled data point $(X, Y)$ in an evaluation set, we observe the abstention decision $R = \sfr(\pi(X), \xi)$ for some independent source of randomness $\xi$ used by the abstaining classifier. 
Then, its prediction $f(X)$ is observed by the evaluator if and only if $R=0$.
Let $S := \sfs(f(X), Y)$ denote the score of the prediction $f$ on the query $X$, irrespective of $R$.
Because $S$ is not observable when $R = 1$, we refer to $S$ as the \emph{potential score} that \emph{would have been seen} had the classifier not abstained.
(See Appendix~\ref{sec:potential_outcomes} for equivalent formulations that explicitly invoke \citet{rubin1974estimating}'s potential outcomes model.)
\iftoggle{compact} {
    \begin{wraptable}{r}{0.55\textwidth}
        \vspace{-0.5\baselineskip}
        \caption{A summary of problem formulations and proposed approaches for evaluation and comparison of abstaining classifiers. 
        Our approaches avoid parametric assumptions and allow for black-box classifiers.} 
        \vspace{-0.5\baselineskip}
        \label{tbl:formulation}
        \begin{center}
        \small
        \begin{tabular}{ccc}
            \toprule 
             & \bf Evaluation & \bf Comparison \\ \midrule
            Classifier(s) & $(f, \pi)$ & $(f^\sfA, \pi^\sfA)$ \& $(f^\sfB, \pi^\sfB)$ \\ 
            Target & $\psi = \ex{S}$ & $\Delta^{\sfA\sfB} = \ex{S^\sfA - S^\sfB}$ \\ \midrule
            Identification & \multicolumn{2}{c}{MAR \& positivity} \\ 
            Estimation & \multicolumn{2}{c}{Doubly robust CI} \\ 
            Optimality & \multicolumn{2}{c}{Nonparametrically efficient} \\
            \bottomrule
        \end{tabular}
        \end{center}
    \vspace{-\baselineskip}
    \end{wraptable}
}{
\begin{table}[t]
    \centering
    \caption{Problem formulation for the counterfactual evaluation of black-box abstaining classifiers. 
    The proposed approaches are applicable to settings where the evaluator has no access to the underlying abstention mechanisms or base predictors, and they do not rely on parametric modeling assumptions.} 
    \label{tbl:formulation}
    \begin{tabular}{ccc}
        \toprule 
         & \bf Evaluation & \bf Comparison \\ \midrule
        Classifier(s) & $(f, \pi)$ & $(f^\sfA, \pi^\sfA)$ \& $(f^\sfB, \pi^\sfB)$ \\ 
        Target & $\psi = \ex{S}$ & $\Delta^{\sfA\sfB} = \ex{S^\sfA - S^\sfB}$ \\ \midrule
        Identification & \multicolumn{2}{c}{MAR \& positivity} \\ 
        Estimation & \multicolumn{2}{c}{Doubly robust CI} \\ 
        Optimality & \multicolumn{2}{c}{Nonparametrically efficient} \\
        \bottomrule
    \end{tabular}
\end{table}
}
Since our evaluation is based only on the score $S$, we can suppress the role of $Y$ and assume that $S$ is observed directly when $R = 0$. 
Similarly, we can suppress the role of $\xi$, which is independent of the data.
We let $\Psymb$ denote the law of $Z:= (X, R, S)$.

Table~\ref{tbl:formulation} summarizes the problem formulations and the proposed approaches in our paper. 
The target definitions and the identification conditions are discussed in the following subsections; the estimator and its optimality are discussed in Section~\ref{sec:estimation_cfscore}.

\subsection{The counterfactual score}\label{sec:target_cfscore}

We propose to assess an abstaining classifier $(f,\pi)$ with its \emph{(expected) counterfactual score}: 
\begin{equation}\label{eqn:cf_score}
    \psi := \ex{S},
\end{equation}
where the expectation is taken w.r.t.\ $\Psymb$.
In words, $\psi$ refers to the expected score of the abstaining classifier had it not been given the option to abstain.
The counterfactual score captures the performance of an abstaining classifier via the score of its base classifier, making it suitable for cases where the evaluator is interested in the predictions without using an abstention mechanism. 

Note that $\psi$ does \emph{not} in general equal the \emph{selective score}, i.e., $\condex{S}{R=0}$. 
For example, when a classifier abstains from making predictions on its ``weak points,'' i.e., inputs on which the classifier performs poorly, the counterfactual score will be lower than the selective score.
Also see Appendix~\ref{sec:condessa} for a direct comparison with \citet{condessa2017performance}'s score\newlyadded{, which \emph{rewards} having bad hidden predictions, as opposed to~\eqref{eqn:cf_score}. 
Our general framework allows estimating either score, although we focus on~\eqref{eqn:cf_score} based on our motivating examples (Example~\ref{ex:api} and Appendix~\ref{sec:examples})}.

\paragraph{Comparison} Counterfactual scores may also be used to compare two abstaining classifiers, $(f^\sfA, \pi^\sfA)$ and $(f^\sfB, \pi^\sfB)$, in the form of their \emph{counterfactual score difference}: $\Delta^{\sfA\sfB} := \psi^{\sfA} - \psi^{\sfB} = \mathbb{E}[S^{\sfA} - S^{\sfB}]$.
Here, the expectation is now taken over the joint law of $Z^{\sfA\sfB} := (X, R^{\sfA}, S^{\sfA}, R^{\sfB}, S^{\sfB})$.

\subsection{Identifiability of the counterfactual score}\label{sec:identification_cfscore}

Having defined the target parameters $\psi$ and $\Delta^{\sfA\sfB}$, we now discuss the assumptions under which these quantities become identifiable using only the observed random variables. In other words, these assumptions establish when the counterfactual quantity equals a statistical quantity.
As in standard settings of counterfactual inference under missing data, the identifiability of counterfactual scores in this setting depends on two standard conditions: (i) the missing at random condition and (ii) positivity.

The \emph{missing at random (MAR)} condition, also known as the \emph{ignorability} or \emph{no unmeasured confounding} condition, requires that the score $S$ is conditionally independent of the abstention decision $R$ given $X$, meaning that there are no unobserved confounders $U$ that affect both the abstention decision $R$ as well as the score $S$. 
Note that $S$ is the \emph{potential} score of what the classifier would get had it not abstained --- it is only observed when $R=0$.
We formally state the MAR condition as follows: 
\begin{assumption}[Scores are missing at random]\label{assn:mar_cfscore} 
$S \indep R \mid X$.
\end{assumption}

In standard ML evaluation scenarios, where the evaluation set is independent of the training set for the classifier, Assumption~\ref{assn:mar_cfscore} is always met. 
We formalize this sufficient condition for MAR in the following proposition.
Let $\calD_{\mathrm{train}}$ denote the collection of any training data used to learn the abstaining classifier $(f, \pi)$ and, as before, $(X, Y)$ denote an (i.i.d.) data point in the evaluation set. 
\begin{proposition}[Independent evaluation data ensures MAR]\label{ppn:eval_indep_mar}
If $(X, Y) \indep \calD_{\mathrm{train}}$, then $S \indep R \mid X$.
\end{proposition}
This result is intuitive: given an independent test input $X$, the score $S = \sfs(f(X), Y)$ is a deterministic function of the test label $Y$, and the abstention decision $R$ of a classifier cannot depend on $Y$ simply because the classifier has no access to it.
A short proof is given in Appendix~\ref{sec:proof_eval_indep_mar} for completeness.
In Appendix~\ref{sec:mar_graph}, we also include causal graphs that visually illustrate how the MAR condition is met.

If the evaluation data is not independent of $\calD_{\mathrm{train}}$, then the classifier already has information about the data on which it is tested, so generally speaking, no evaluation score will not accurately reflect its generalization performance.
Although the independence between the training and evaluation data is expected in standard ML applications, it may not be guaranteed when, e.g., using a publicly available dataset that is used during the training of the classifier. 
These issues can be prevented by ensuring that the evaluation set is held out (e.g., a hospital can use its own patient data to evaluate APIs).

The second condition, the \emph{positivity} condition, is more substantial in our setting: 
\begin{assumption}[Positivity]\label{assn:positivity_cfscore}
There exists $\epsilon > 0$ such that $\pi(X) = \condprob{R=1}{X} \leq 1 - \epsilon$.
\end{assumption}
Assumption~\ref{assn:positivity_cfscore} says that, for each input $X$, there has to be at least a small probability that the classifier will \emph{not} abstain ($R=0$).
Indeed, if the classifier deterministically abstains from making predictions on a specific input that has nonzero marginal density, then we have no hope of estimating an expectation over all possible values that $X$ can take. 
When it comes to evaluating abstaining classifiers on safety-critical applications, we argue that this condition may need to be enforced at a policy level --- we elaborate on this point in Section~\ref{sec:discussion} and Appendix~\ref{sec:policy}.
In practice, the exact value of $\epsilon$ is problem-dependent, and in Appendix~\ref{sec:positivity_experiment}, we include additional experiments illustrating how our methods retain validity as long as the abstention rate is capped at $1-\epsilon$ for some $\epsilon>0$.

Another justification for the positivity condition is that stochastically abstaining classifiers can achieve better performances than their deterministic counterparts.
\citet{kalai2021towards} illustrate how stochastic abstentions can improve the out-of-distribution (OOD) performance w.r.t. the \citet{chow1970optimum} score (i.e., $\mathrm{selective\;score} + \alpha \cdot \mathrm{coverage}$).
\citet{schreuder2021classification} also introduce randomness in their abstaining classifiers, which leverage abstentions as a means to improve their accuracy while satisfying a fairness constraint.
The role of random abstentions in these examples mirrors the role of randomization in the fairness literature~\citep{barocas2019fairness}, where the optimal randomized fair predictors are known to outperform their deterministic counterparts~\citep{agarwal2022power,grgichlaca2017fairness}.
Given the effectiveness of randomized classifiers for fairness, it would not be surprising if a fair abstaining classifier was randomized (in its decisions and abstentions).

With MAR and positivity in hand, we can show that the counterfactual score is indeed identifiable.
Define $\mu_0(x) := \ex{S \mid R=0, X=x}$ as the regression function for the score under $R=0$.
\begin{proposition}[Identification]\label{ppn:identification_cfscore}
Under Assumptions~\ref{assn:mar_cfscore} and~\ref{assn:positivity_cfscore}, $\psi$ is identified as $\ex{\mu_0(X)}$.
\end{proposition}
The proof, included in Appendix~\ref{sec:proof_identification_cfscore}, follows a standard argument in causal inference.
The identification of the target parameter $\psi$ using $\mu_0$ implies that we can estimate $\psi$, the expectation of a potential outcome, using only the observed inputs and scores.
Specifically, the task of estimating $\psi$ consistently reduces to the problem of estimating the regression function $\mu_0$, which only involves predictions that the classifier did not abstain from making.
We note that, as in standard causal inference, the task of identification, which concerns \emph{what} to estimate, is largely orthogonal to the task of estimation, which concerns \emph{how} to estimate the quantity.
We discuss the latter problem in Section~\ref{sec:estimation_cfscore}.

\paragraph{Comparison} For the comparison task, given $\Delta^{\sfA\sfB} = \psi^{\sfA} - \psi^{\sfB}$, it immediately follows that if the MAR and positivity assumptions hold for each of $(X, R^{\sfA}, S^{\sfA})$ and $(X, R^{\sfB}, S^{\sfB}),$ then $\Delta^{\sfA\sfB}$ is also identified as $\Delta^{\sfA\sfB} = \mathbb{E}[\mu^\sfA_0(X) - \mu^\sfB_0(X)]$,
where $\mu^\bullet_0(x) := \ex{S^{\bullet}|R^{\bullet} = 0,X=x}$ for $\bullet \in \{\sfA, \sfB\}$.
In words, if (i) the evaluation data is independent of the training data for each classifier (Proposition~\ref{ppn:eval_indep_mar}) and (ii) each classifier has at least a small chance of not abstaining on each input (Assumption~\ref{assn:positivity_cfscore}), then the counterfactual score difference, i.e., $\Delta^{\sfA\sfB}$, is identified as the expected difference in the expected scores over inputs conditional on non-abstentions, i.e., $\mathbb{E}[\mu^\sfA_0(X) - \mu^\sfB_0(X)]$.

\section{Nonparametric and doubly robust estimation of the counterfactual score}\label{sec:estimation_cfscore}

Having identified the counterfactual scores, we now focus on the problem of consistently estimating them.
We estimate these quantities without resorting to parametric assumptions about the underlying black-box abstention mechanisms.
Instead, we reduce the problem to that of functional estimation and leverage techniques from nonparametric statistics. 
See \citet{kennedy2022semiparametric} for a recent review.

\subsection{Estimating the counterfactual score}\label{sec:estimation_cfscore_evaluation}

\paragraph{Task} Let $\{(X_i, R_i, S_i)\}_{i=1}^n \sim \Psymb$ denote an i.i.d.~evaluation set of size $n$.
As before, we assume that we are given access to the censored version of this sample, i.e., that we observe $S_i$ if and only if $R_i = 0$. 
Using the observables, we seek to form an estimate $\hat\psi$ of the counterfactual score $\psi = \mathbb{E}[S]$. 

\paragraph{Doubly robust estimation} 
Under identification (Proposition~\ref{ppn:identification_cfscore}), we can estimate $\psi$ by estimating the regression function $\mu_0(X)$ on the data $\{(X_i,S_i): R_i = 0\}$. However, the na\"{i}ve ``plug-in'' estimate suffers from an inflated bias due to the structure present in the abstention patterns. (See Appendix~\ref{sec:baselines} for details.)
We instead develop a doubly robust (DR) estimator~\citep{robins1994estimation,bang2005doubly}, which is known to consistently estimate $\psi$ at the optimal \emph{nonparametric efficiency rates}, meaning that no other estimator based on the $n$ observations can asymptotically achieve a smaller mean squared error \citep{vandervaart2002semiparametric}. The derivation below is relatively standard, explaining our brevity.

Formally, the DR estimator is defined using the (uncentered) \emph{efficient influence function (EIF)} for the identified target functional $\psi(\Psymb) = \mathbb{E}_\Psymb[\mu_0(X)]$: $\influence(x, r, s) := \mu_0(x) + \frac{1-r}{1-\pi(x)}\inparen{s - \mu_0(x)}$ ($0/0:=0$).
Here, $\pi$ and $\mu_0$ are the ``nuisance'' functions, representing the abstention mechanism and the score function under $R=0$, respectively. 
The EIF can be computed as long as $s$ is available when $r=0$.
An intuition for the EIF is that it is the first-order ``distributional Taylor approximation''~\citep{fisher2021visually} of the target functional, such that its bias is second-order.

Given that $\pi$ and $\mu_0$ are unknown, we define an estimate of the EIF, denoted as $\hat\influence{}$, by plugging in estimates $\hat\pi$ for $\pi$ and $\hat\mu_0$ for $\mu_0$. 
Then, the DR estimator is simply the empirical mean of the EIF:
\begin{equation}\label{eqn:dr}
\hat\psi_\dr = \frac{1}{n} \sum_{i=1}^n \hat\influence(X_i, R_i, S_i) = \frac{1}{n} \sum_{i=1}^n \insquare{ \hat\mu_0(X_i) + \frac{1-R_i}{1-\hat\pi(X_i)}\inparen{S_i - \hat\mu_0(X_i)}}.
\end{equation}
This estimator is well-defined because $S_i$ is available precisely when $R_i = 0$.
Note that the first term is the (biased) plug-in estimator, and the second term represents the first-order correction term, which involves inverse probability weighting (IPW)~\citep{horvitz1952generalization,rosenbaum1995design}.
In our experiments, we show how the DR estimator improves upon both the plug-in estimator, in terms of the bias, and the IPW-based estimator, which we recap in Section~\ref{sec:baselines}, in terms of the variance.

The ``double robustness'' of $\hat\psi_\dr$ translates to the following useful property: $\hat\psi_\dr$ retains the parametric rate of convergence, $O_\Psymb(1/\sqrt{n})$, even when the estimators $\hat\mu_0$ and $\hat\pi$ themselves converge at slower rates.
This allows us to use nonparametric function estimators to estimate $\mu_0$ and $\pi$, such as stacking ensembles~\citep{breiman1996stacked} like the super learner~\citep{vanderlaan2007super} and regularized estimators like the Lasso~\citep{tibshirani1996regression,belloni2014inference}. 
Even for nonparametric models whose rates of convergence are not fully understood, such as random forests~\citep{breiman2001randomforests} and deep neural networks~\citep{lecun2015deep}, we can empirically demonstrate valid coverage and efficiency (Section~\ref{sec:experiments}).

In practice, the nuisance functions can be estimated via \emph{cross-fitting}~\citep{robins2008higher,zheng2011crossvalidated,chernozhukov2018double}, which is a $K$-fold generalization of sample splitting.
First, randomly split the data into $K$ folds; then, fit $\hat\pi$ and $\hat\mu_0$ on $K-1$ folds and use them to estimate the EIF on the remaining ``evaluation'' fold; repeat the process $K$ times with each fold being the evaluation fold; finally, average the EIFs across all data points.
The key benefit of using cross-fitting is to avoid any complexity restrictions on individual nuisance functions without sacrificing sample efficiency.
In the following, we let $\hat\psi_\dr$ be the estimator~\eqref{eqn:dr} obtained via cross-fitting.

Now we are ready to present our first result, which states the asymptotic validity and efficiency of the DR estimator for $\psi$ under identification and the DR condition.
\begin{theorem}[DR estimation of the counterfactual score for an abstaining classifier]\label{thm:evaluation_cfscore}
Suppose that Assumptions~\ref{assn:mar_cfscore} and~\ref{assn:positivity_cfscore} hold.
Also, suppose that
\begin{equation}\label{eqn:dr_assn}
    \norm{\hat\pi - \pi}_{L_2(\Psymb)}\norm{\hat\mu_0 - \mu_0}_{L_2(\Psymb)} = o_\Psymb(1/\sqrt{n})
\end{equation}
and that $\lVert\hat\influence{} - \influence\rVert_{L_2(\Psymb)} = o_\Psymb(1)$. 
Then, 
\[
    \sqrt{n}\inparen{\hat\psi_\dr - \psi} \rightsquigarrow \calN\inparen{0, \Var{\Psymb}{\influence}},
\]
where $\Var{\Psymb}{\influence}$ matches the nonparametric efficiency bound.
\end{theorem}
The proof adapts standard arguments in mathematical statistics, as found in, e.g., \citet{vandervaart2002semiparametric,kennedy2022semiparametric}, to the abstaining classifier evaluation setup.
We include a proof sketch in Appendix~\ref{sec:proof_evaluation_cfscore}.
Theorem~\ref{thm:evaluation_cfscore} tells us that, under the identification and the DR condition~\eqref{eqn:dr_assn}, we can construct a closed-form asymptotic confidence interval (CI) at level $\alpha \in (0,1)$ as follows:
\begin{equation}\label{eqn:ci}
    C_{n,\alpha} = \inparen{ \hat\psi_\dr\pm z_{\alpha/2} \sqrt{n^{-1}\Var{\hat\Psymb_n}{\hat\influence{}}} },
\end{equation}
where $z_{\alpha/2} = \Phi(1 - \frac{\alpha}{2})$ is the $(1- \frac{\alpha}{2})$-quantile of a standard normal (e.g., $1.96$ for $\alpha=0.05$).
In Appendix~\ref{sec:asympcs}, we describe an asymptotic confidence sequence~\citep[AsympCS;][]{waudbysmith2021doubly}, which is a CI that is further valid under continuous monitoring (e.g., as more data is collected).

\subsection{Estimating counterfactual score differences}\label{sec:estimation_cfscore_comparison}

\paragraph{Task} Next, we return to the problem of comparing two abstaining classifiers, $(f^\sfA, \pi^\sfA)$ and $(f^\sfB, \pi^\sfB)$, that each makes a decision to make a prediction on each input $X_i$ or abstain from doing so.
That is, $R^{\bullet}_i \mid \pi^\bullet(X_i) \sim \mathsf{Ber}(\pi^\bullet(X_i))$, and we observe $S^{\bullet}_i = \sfs(f^\bullet(X_i), Y_i)$ if and only if $R^{\bullet}_i=0$, for $\bullet \in \{\sfA, \sfB\}$. 
Recall that the target here is the score difference $\Delta^{\sfA\sfB} = \psi^{\sfA} -\psi^{\sfB} = \ex{S^{\sfA} - S^{\sfB}}$.

\paragraph{Doubly robust difference estimation} 
If the parameters $\psi^{\sfA}$ and $\psi^{\sfB}$ are each identified according to Proposition~\ref{ppn:identification_cfscore}, then we can estimate $\Delta^{\sfA\sfB}$ as $\hat\Delta^{\sfA\sfB} = \hat\psi^{\sfA} - \hat\psi^{\sfB},$ for individual estimates $\hat\psi^{\sfA}$ and $\hat\psi^{\sfB}$. 
The resulting EIF is simply the difference in the EIF for $\sfA$ and $\sfB$: $\influence^{\sfA\sfB}(x, r^\sfA, r^\sfB, s^\sfA, s^\sfB) = \influence^\sfA(x, r^\sfA, s^\sfA) - \influence^\sfB(x, r^\sfB, s^\sfB)$, where $\influence^\sfA$ and $\influence^\sfB$ denote the EIF of the respective classifier.
Thus, we arrive at an analogous theorem that involves estimating the nuisance functions of each abstaining classifier and utilizing $\influence^{\sfA\sfB}$ to obtain the limiting distribution of $\hat\Delta_\dr^{\sfA\sfB} = \hat\psi_\dr^\sfA - \hat\psi_\dr^\sfB$.
\begin{theorem}[DR estimation of the counterfactual score difference]\label{thm:comparison_cfscore}
Suppose that Assumptions~\ref{assn:mar_cfscore} and~\ref{assn:positivity_cfscore} hold for both $(X_i, R^{\sfA}_i, S^{\sfA}_i)$ and $(X_i, R^{\sfB}_i, S^{\sfB}_i)$.
Also, suppose that
\begin{align}
    \lVert \hat\pi^\sfA - \pi^\sfA \rVert_{L_2(\Psymb)} \lVert \hat\mu_0^{ \sfA} - \mu_0^{ \sfA} \rVert_{L_2(\Psymb)} + \lVert \hat\pi^\sfB - \pi^\sfB \rVert_{L_2(\Psymb)} \lVert\hat\mu_0^{ \sfB} - \mu_0^{ \sfB} \rVert_{L_2(\Psymb)} = o_\Psymb(1/\sqrt{n})  \label{eqn:dr_assn_comparison}
\end{align}
and that $\lVert\hat\influence{}^{\sfA\sfB} - \influence^{\sfA\sfB}\rVert_{L_2(\Psymb)} = o_\Psymb(1)$.
Then, 
\[
    \sqrt{n}\inparen{\hat\Delta_\dr^{\sfA\sfB} - \Delta^{\sfA\sfB}} \rightsquigarrow \calN\inparen{0, \Var{\Psymb}{\influence^{\sfA\sfB}}},
\]
where $\Vsymb_{\Psymb}[\influence^{\sfA\sfB}]$ matches the nonparametric efficiency bound.
\end{theorem}
A proof is given in Appendix~\ref{sec:proof_comparison_cfscore}. 
As with evaluation, 
Theorem~\ref{thm:comparison_cfscore} yields a closed-form asymptotic CI of the form~\eqref{eqn:ci} using the analogous estimate of EIF under MAR, positivity, and DR~\eqref{eqn:dr_assn_comparison}. 
Inverting this CI further yields a hypothesis test for $H_0 : \psi^\sfA = \psi^\sfB$ vs. $H_1: \psi^\sfA \neq \psi^\sfB$.


\section{Experiments}\label{sec:experiments}

The main goals of our experiments are to empirically check the validity and power of our estimation methods and to illustrate how our methods can be used in practice.
First, in Section~\ref{sec:simulated}, we present results on simulated data to examine the validity of our proposed inference methods (CIs and hypothesis tests). 
Then, in Section~\ref{sec:cifar100}, we study three scenarios on the CIFAR-100 dataset that illustrate the practical use of our approach to real data settings. 
All code for the experiments is publicly available online at \url{https://github.com/yjchoe/ComparingAbstainingClassifiers}.

\subsection{Simulated experiments: Abstentions near the decision boundary}\label{sec:simulated}

\paragraph{Setup (MAR but not MCAR)} We first consider comparing two abstaining classifiers according to their accuracy scores, on a simulated binary classification dataset with 2-dimensional inputs.
Given $n=2,000$ i.i.d.\ inputs $\{X_i\}_{i=1}^n \sim \mathsf{Unif}([0,1]^2)$, each label $Y_i$ is decided using a linear boundary, $f_*(x_1, x_2) = \indicator{x_1 + x_2 \geq 1}$, along with a 15\% i.i.d.\ label noise.
Importantly, each classifier abstains near its decision boundary, such that its predictions and scores are \emph{MAR but not MCAR} (because abstentions depend on the inputs).
As a result, while the counterfactual score of $\sfA$ ($\psi^\sfA = 0.86$) is much higher than $\sfB$ ($\psi^\sfB = 0.74$), their selective scores are more similar ($\mathsf{Sel}^\sfA = 0.86$, $\mathsf{Sel}^\sfB = 0.81$) and coverage is lower for $\sfA$ ($\mathsf{Cov}^\sfA = 0.55$) than for $\sfB$ ($\mathsf{Cov}^\sfB = 0.62$).
Another point to note here is that, even though both the outcome model and classifier $\sfA$ are linear, both the abstention mechanism\footnote{The abstention mechanism $\pi(x) = \mathbb{P}(R=1\mid X=x)$ here separates the region below \emph{and} above the decision boundary from the region near the boundary. Thus $\pi$ is nonlinear even when the boundary is linear.} $\pi^\sfA$ and the selective score function\footnote{Given any input $X$ for which the classifier did not abstain ($R=0$) and its output $Y$, the score $S = \sfs(f(X), Y)$ is nonlinear if either $\sfs(\cdot, y)$ or $f$ is nonlinear. Thus, even for linear $f$, nonlinear scores like the Brier score automatically make the selective score function $\mu_0(x) = \mathbb{E}[S \mid R=0, X=x]$ nonlinear.}  $\mu_0^\sfA$ are \emph{non}linear functions of the inputs (similarly for $\pi^\sfB$ and $\mu_0^\sfB$).
More generally, if a classifier abstains near its decision boundary, then both $\pi$ and $\mu_0$ could be at least as complex as the base classifier $f$ itself.
Further details of the setup, including a plot of the data, predictions, and abstentions, are provided in Appendix~\ref{sec:binary_mar_setup}.

\iftoggle{compact}{
    \begin{wraptable}{R}{0.55\textwidth}
    \vspace{-0.5\baselineskip}
    \caption{Miscoverage rates (and widths) of 95\% CIs using three estimation approaches and three nuisance function ($\pi$ and $\mu_0$) estimators in a simulated experiment. 
    Mean and standard error computed over $m=1,000$ runs are shown; those within 2 standard errors of the intended level ($0.05$) are boldfaced.
    The sample size is $n=2,000$ in each run.
    The mean widths of CIs are shown in parentheses.
    DR estimation with either a random forest or a super learner achieves control over the miscoverage rate, and the DR-based CI is twice as tight as the IPW-based CI in terms of their width.} 
    \vspace{-0.5\baselineskip}
    \begin{center}
    \begin{small}
    \begin{tabular}{lccc}
    \toprule
    \bf $\hat\mu_0$ \& $\hat\pi$ & \bf Plug-in & \bf IPW & \bf DR \\
    \midrule
    Linear & 1.00 $\pm$ 0.00 & 0.76 $\pm$ 0.01 & 1.00 $\pm$ 0.00 \\
           & (0.00)		  & (0.09)		    & (0.04)		 \\  \midrule
    Random & 0.64 $\pm$ 0.02 & 0.14 $\pm$ 0.01 & \bf 0.05 $\pm$ 0.01 \\
    forest & (0.02)		  & (0.13)		    & \bf (0.07)		 \\ \midrule
    Super  & 0.91 $\pm$ 0.01 & \bf 0.03 $\pm$ 0.01 & \bf 0.05 $\pm$ 0.01 \\
    learner & (0.01)		  & \bf (0.12)		    & \bf (0.06)		 \\
    \bottomrule
    \end{tabular}
    \end{small}
    \end{center}
    \vspace{-2\baselineskip}
    \label{tbl:miscoverage_bias}
    \end{wraptable}
}{
    \begin{table}[t]
    \centering
    \caption{Miscoverage rates (and widths) of 95\% CIs using three estimation approaches and three nuisance function ($\pi$ and $\mu_0$) estimators in a simulated experiment. 
    Mean and standard error computed over $m=1,000$ runs are shown; those within 2 standard errors of the intended level ($0.05$) are boldfaced.
    The sample size is $n=2,000$ in each run.
    The mean widths of CIs are shown in parentheses.
    DR estimation with either a random forest or a super learner achieves control over the miscoverage rate, and the DR-based CI is twice as tight as the IPW-based CI in terms of their width.} 
    \label{tbl:miscoverage_bias}
    \begin{tabular}{lccc}
    \toprule
    \bf Nuisance Function Estimators & \bf Plug-in & \bf IPW & \bf DR \\
    \midrule
    Linear/Logistic Regression & 1.00 $\pm$ 0.00 & 0.76 $\pm$ 0.01 & 1.00 $\pm$ 0.00 \\
           & (0.00)		  & (0.09)		    & (0.04)		 \\  \midrule
    Random Forest & 0.64 $\pm$ 0.02 & 0.14 $\pm$ 0.01 & \bf 0.05 $\pm$ 0.01 \\
           & (0.02)		  & (0.13)		    & \bf (0.07)		 \\ \midrule
    Super Learner  & 0.91 $\pm$ 0.01 & \bf 0.03 $\pm$ 0.01 & \bf 0.05 $\pm$ 0.01 \\
           & (0.01)		  & \bf (0.12)		    & \bf (0.06)		 \\
    \bottomrule
    \end{tabular}
    \end{table}
}

\paragraph{Miscoverage rates and widths}
As our first experiment, we compare the miscoverage rates and widths of the 95\% DR CIs (Theorem~\ref{thm:comparison_cfscore}) against two baseline estimators: the plug-in and the IPW \citep{rosenbaum1995design} estimators (Appendix~\ref{sec:baselines}).
For each method, the miscoverage rate of the CI $C_n$ is approximated via $\Psymb(\Delta^{\sfA\sfB} \notin C_n) \approx m^{-1}\sum_{j=1}^m \mathbf{1}(\Delta^{\sfA\sfB} \notin C_n^{(j)})$, where $m$ is the number of simulations over repeatedly sampled data.
If the CI is valid, then this rate should approximately be $0.05$.
The miscoverage rate and the width of a CI, respectively, capture its bias and variance components.
For the nuisance functions, we try linear predictors (L2-regularized linear/logistic regression for $\hat\mu_0$/$\hat\pi$), random forests, and super learners with $k$-NN, kernel SVM, and random forests.

We present our results in Table~\ref{tbl:miscoverage_bias}.
First, using the random forest or the super learner, the DR CIs consistently achieve the intended coverage level of $0.95$, over $m=1,000$ repeated simulations (standard error $0.01$).
This validates the asymptotic normality result of~\eqref{thm:comparison_cfscore}.
Note that the CI with linear estimators does not achieve the intended coverage level: this is expected as neither $\hat\pi$ nor $\hat\mu_0$ can consistently estimate the nonlinear functions $\pi$ or $\mu_0$, violating the DR condition~\eqref{eqn:dr_assn_comparison}.

Second, when considering both the miscoverage rate and CI width, the DR estimator outperforms both the plug-in and IPW estimators.
The plug-in estimator, despite having a small CI width, has a very high miscoverage rate ($0.91$ with the super learner), meaning that it is biased even when flexible nuisance learners are used.
On the other hand, the IPW estimator has double the width of the DR estimator ($0.12$ to $0.06$, with the super learner), meaning that it is not as efficient. 
Also, while the IPW estimator achieves the desired coverage level with the super learner ($0.03$), it fails with the random forest ($0.14$), which tends to make overconfident predictions of the abstention pattern and biases the resulting estimate.
In contrast, the DR estimator retains its intended coverage level of $0.05$ with the random forest, suggesting that it is amenable to overconfident nuisance learners.

\paragraph{Power analysis}
We conduct a power analysis of the statistical test for $H_0: \Delta^{\sfA\sfB} = 0$ vs. $H_1: \Delta^{\sfA\sfB} \neq 0$ by inverting the DR CI.
The results confirm that the power reaches 1 as either the sample size ($n$) or the absolute difference ($|\Delta^{\sfA\sfB}|$) increases.
This experiment is included in Appendix~\ref{sec:power}.

\subsection{Comparing abstaining classifiers on CIFAR-100}\label{sec:cifar100}

To illustrate a real data use case, we compare abstaining classifiers on the CIFAR-100 image classification dataset~\citep{krizhevsky2009learning}.
Observe that abstaining classifiers can behave differently not just when their base classifiers are different but also when their abstention mechanisms are different.
In fact, two abstaining classifiers can have a similarly effective base classifier but substantially different abstention mechanisms (e.g., one more confident than the other). 
In such a case, the counterfactual score difference between the two classifiers is zero, but their selective scores and coverages are different. 
We examine such scenarios by comparing image classifiers that use the same pre-trained representation model but have different output layers and abstention mechanisms.

We start with the 512-dimensional final-layer representations of a VGG-16 convolutional neural network~\citep{simonyan2015very}, pre-trained\footnote{Reproduced version, accessed from \url{https://github.com/chenyaofo/pytorch-cifar-models}.} on the CIFAR-100 training set, and compare different output layers and abstention mechanisms on the validation set.
Generally, in a real data setup, we cannot verify whether a statistical test or a CI is correct; yet, in this experiment, we can still access to the base model of each abstaining classifier.
This means that (a) if we compare abstaining classifiers that share the base classifier but differ in their abstention patterns, then we actually know that their counterfactual scores are exactly the same ($\Delta^{\sfA\sfB} = 0$); (b) if we compare abstaining classifiers with different base classifiers, then we can compute their counterfactual scores accurately up to an i.i.d.~sampling error. 
This estimate is denoted by $\bar\Delta^{\sfA\sfB} := n^{-1}\sum_{i=1}^n [\sfs(f^\sfA(X_i), Y_i) - \sfs(f^\sfB(X_i), Y_i)]$. 

For all comparisons, we use the DR estimator, where the nuisance functions $\hat\pi$ and $\hat\mu_0$ for both classifiers are each an L2-regularized linear layer learned on top of the pre-trained VGG-16 features. 
{The use of pre-trained representations for learning the nuisance functions is motivated by \citet{shi2019adapting}, who demonstrated the effectiveness of the approach in causal inference contexts.}
We also use the Brier score in this experiment.
We defer other experiment details to Appendix~\ref{sec:cifar100_details}.

\iftoggle{compact}{
    \begin{wraptable}{r}{0.55\textwidth}
    \vspace{-0.5\baselineskip}
    \caption{The 95\% DR CIs and their corresponding hypothesis tests for $H_0: \Delta^{\sfA\sfB} = 0$ at significance level $\alpha=0.05$, for three different comparison scenarios on (half of) the CIFAR-100 test set ($n=5,000$).
    The three scenarios compare different abstention mechanisms or predictors, as detailed in text; all comparisons use the Brier score.
    $\bar\Delta^{\sfA\sfB}$ is the empirical counterfactual score difference without any abstentions. 
    The result of each statistical test agrees with whether $\bar\Delta^{\sfA\sfB}$ is $0$.}
    \vspace{-0.5\baselineskip}
    \begin{center}
    \begin{small}
    \begin{tabular}{crcc}
    \toprule
    \bf Scenarios & \bf $\bar\Delta^{\sfA\sfB}$ & \bf 95\% DR CI & \bf Reject $H_0$? \\
    \midrule
    \texttt{I}   & $0.000$  & (-0.005, 0.018) & No  \\
    \texttt{II}  & $0.000$  & (-0.014, 0.008) & No  \\
    \texttt{III} & $-0.029$ & (-0.051, -0.028) & Yes   \\
    \bottomrule
    \end{tabular}
    \end{small}
    \end{center} 
    \vspace{-\baselineskip}
    \label{tbl:cifar100}
    \end{wraptable}
}{
    \begin{table}[t]
    \centering
    \caption{The 95\% DR CIs and their corresponding hypothesis tests for $H_0: \Delta^{\sfA\sfB} = 0$ at significance level $\alpha=0.05$, for three different comparison scenarios on (half of) the CIFAR-100 test set ($n=5,000$).
    The three scenarios compare different abstention mechanisms or predictors, as detailed in text; all comparisons use the Brier score.
    $\bar\Delta^{\sfA\sfB}$ is the empirical counterfactual score difference without any abstentions. 
    The result of each statistical test agrees with whether $\bar\Delta^{\sfA\sfB}$ is $0$.}    
    \label{tbl:cifar100}
    \begin{tabular}{crcc}
    \toprule
    \bf Scenarios & \bf $\bar\Delta^{\sfA\sfB}$ & \bf 95\% DR CI & \bf Reject $H_0$? \\
    \midrule
    \texttt{I}   & $0.000$  & (-0.005, 0.018) & No  \\
    \texttt{II}  & $0.000$  & (-0.014, 0.008) & No  \\
    \texttt{III} & $-0.029$ & (-0.051, -0.028) & Yes   \\
    \bottomrule
    \end{tabular}
    \end{table}
}

In scenario \texttt{I}, we compare two abstaining classifiers that use the same softmax output layer but use a different threshold for abstentions.
Specifically, both classifiers use the softmax response (SR) thresholding~\citep{geifman2017selective}, i.e., abstain if $\max_{c\in \calY}f(X)_c < \tau$ for a threshold $\tau>0$, but $\sfA$ uses a more conservative threshold ($\tau=0.8$) than $\sfB$ ($\tau=0.5$). 
As a result, while their counterfactual scores are identical ($\Delta^{\sfA\sfB}=0$), $\sfA$ has a higher selective score ($+0.06$) and a lower coverage ($-0.20$) than $\sfB$.
This is also a deterministic abstention mechanism, potentially challenging the premises of our setup.
As shown in Table~\ref{tbl:cifar100}, we see that the 95\% DR CI is $(-0.005, 0.018)$ ($n=5,000$), confirming that there is no difference in counterfactual scores.\footnote{The reason why the DR CI correctly estimates the true $\Delta^{\sfA\sfB}=0$, despite the fact that positivity is violated in this case, is because the two classifiers happen to abstain on similar examples ($\sfB$ abstains whenever $\sfA$ does) \emph{and} their scores on their abstentions happen to be relatively similar ($0.604$ for $\sfA$; $0.576$ for $\sfB$).}

Scenario \texttt{II} is similar to scenario \texttt{I}, except that the abstention mechanisms are now stochastic: $\sfA$ uses one minus the SR as the probability of making a prediction, i.e., $\pi^\sfA(x) = 1 - \max_{c\in [C]}f(X)_c$, while $\sfB$ uses one minus the Gini impurity as the probability of abstention, i.e., $\pi^\sfB(x) = 1 - \sum_{c=1}^C f(X)_c^2$, both clipped to $(0.2, 0.8)$.
The Gini impurity is an alternative measure of confidence to SR that is quadratic in the probabilities, instead of piecewise linear, and
thus the Gini-based abstaining classifier ($\sfB$) is more cautious than the SR-based one ($\sfA$), particularly on uncertain predictions.
In our setup, $\sfA$ achieves a higher coverage than $\sfB$, while $\sfB$ achieves a higher selective score than $\sfA$.
The 95\% DR CI is $(-0.014, 0.008)$, confirming that there is once again no difference in counterfactual scores.
Scenarios \texttt{I} and \texttt{II} both correspond to case (a).

In scenario \texttt{III}, we now examine a case where there \emph{is} a difference in counterfactual scores between the two abstaining classifiers (case (b)). 
Specifically, we compare the pre-trained VGG-16 model's output layers (512-512-100) with the single softmax output layer that we considered in earlier scenarios.
It turns out that the original model's multi-layer output model achieves a worse Brier score ($0.758$) than one with a single output layer ($0.787$), likely because the probability predictions of the multi-layer model are too confident (miscalibrated).
When using the same abstention mechanism (stochastic abstentions using SR, as in \texttt{II}), the overconfident original model correspondingly achieves a higher coverage (and worse selective Brier score) than the single-output-layer model.
The Monte Carlo estimate of the true counterfactual score difference is given by $\bar\Delta^{\sfA\sfB} = -0.029$, and the 95\% DR CI falls entirely negative with $(-0.051, -0.028)$, rejecting the null of $\Delta^{\sfA\sfB}=0$ at $\alpha=0.05$.


\section{Discussion}\label{sec:discussion}

This paper lays the groundwork for addressing the challenging problem of counterfactually comparing black-box abstaining classifiers. 
Our solution casts the problem in the missing data framework, in which we treat abstentions as MAR predictions of the classifier(s), and this allows us to leverage nonparametrically efficient tools from causal inference. 
We close with discussions of two topics.

\paragraph{Addressing violations of the identification conditions}
At the conceptual level, the biggest challenge arises from the positivity condition, which requires the classifiers to deploy a non-deterministic abstention mechanism. 
As mentioned in Section~\ref{sec:identification_cfscore}, the counterfactual score is unidentifiable without this assumption.
We argue that this issue calls for a policy-level treatment, especially in auditing scenarios, where the evaluators may require vendors to supply a classifier that can abstain but has at least an $\epsilon>0$ chance of nonabstention.
The level $\epsilon$ can be mutually agreed upon by both parties.
Such a policy achieves a middle ground in which the vendors are not required to fully reveal their proprietary classifiers. 
In Appendix~\ref{sec:policy}, we discuss this policy-level treatment in greater detail. 

An alternative is to explore existing techniques that aim to address positivity violations directly~\citep{petersen2012diagnosing,ju2019adaptive,leger2022causal}.
Due to the unidentifiability result, these techniques have their own limitations. 
For example, we may consider applying sample trimming~\citep{crump2009dealing} to make valid inferences on a subpopulation, but the conclusions would not hold for the entire input domain (even when relevant). 
\newlyadded{Nevertheless, it remains a meaningful future work to adapt modern tools in causal inference for diagnosing and addressing positivity violations~\citep[e.g.,][]{lei2021distribution,kennedy2019nonparametric}.
In the case of addressing MAR violations, if we suspect the evaluation data is ``compromised'' and used in training, we can adapt sensitivity analysis methods that measure how much of the evaluation data is contaminated by the training data~\citep{bonvini2022sensitivity}.}

\paragraph{Connections to learning-to-defer scenarios and cascading classifiers}
There are variants of abstaining classifiers in ML for which we can utilize the counterfactual score analogously.
First, in the learning-to-defer setting~\citep{madras2018predict}, where the algorithm is allowed to defer its decision to an expert that gives their own decision, the counterfactual score naturally corresponds to the expected score of the overall system had the classifier not deferred at all. 
The score is thus an evaluation metric primarily for the classifier, and it is independent of the expert’s predictions, even when the classifier is adaptive to the expert’s tendencies.
In the case where the goal is to assess the \emph{joint} performance of the algorithm and the expert, then a variant of \citet{condessa2017performance}’s score can be useful; in Appendix~\ref{sec:learning_to_defer}, we discuss how the DR estimator can be utilized for estimating this variant, even when the expert's score is random.

Another example would be cascading or multi-stage classifiers~\citep{alpaydin1998cascading,viola2001robust}, which are classifiers that short-circuit easy predictions by using simple models in their early stages to save computation.
The counterfactual framework can assess the performance of cascading classifiers by treating each component as an abstaining classifier. 
For example, a two-stage cascading classifier, equipped with a small model $f_0$, a large model $f_1$, and a deferral mechanism $\pi$, can be viewed as a pair of abstaining classifiers with tied abstention mechanisms, specifically $(f_0,\pi)$ and $(f_1, \bar\pi)$ where $\bar\pi = 1 - \pi$.
Then, the corresponding counterfactual scores capture the performance of the cascading classifier had it been using only that (small or large) model.
We can further extend this to general multi-stage classifiers if we are interested in estimating the score of a classifier in each stage individually.
The two-stage setup is also relevant to the learning-to-defer setup, where the large model is an imperfect expert on the task. 
In such a setup, we can devise a metric that combines (i) the selective score of the small model and (ii) the relative improvement by switching to the expert. 
Our approach can also estimate this new metric using essentially the same tools.

\clearpage 

\subsection*{Acknowledgements}
The authors thank Edward H. Kennedy and anonymous reviewers for their helpful comments and suggestions.
AR acknowledges support from NSF grants IIS-2229881 and DMS-2310718.
This work used the Extreme Science and Engineering Discovery Environment (XSEDE), which is supported by National Science Foundation grant number ACI-1548562. Specifically, it used the Bridges-2 system, which is supported by NSF award number ACI-1928147, at the Pittsburgh Supercomputing Center (PSC).


\bibliography{contents/references}
\bibliographystyle{apalike}

\clearpage

\appendix

\renewcommand\thefigure{App.\arabic{figure}}    
\renewcommand\thetable{App.\arabic{table}}    
\setcounter{figure}{0} 

\section{Further discussion}\label{sec:further}


\subsection{Additional motivating examples for the counterfactual score}\label{sec:examples}

Here, we include three additional examples that motivate the counterfactual score.
These illustrate cases in which either (a) the missing predictions are utilized in a failure mode (Examples~\ref{ex:car} and~\ref{ex:triage}) or (b) the missing predictions are relevant to the evaluator's future uses (Examples~\ref{ex:triage} and~\ref{ex:bias}).

\begin{example}[Inattentive driver in a self-driving car]\label{ex:car}
Consider an ML classifier in a semi-autonomous vehicle system that makes a prediction (the weather, time of day, etc.) given the available sensory inputs.
The prediction is used by the sequential decision making agent.
In principle, when facing a high-uncertainty input, the classifier can abstain from a prediction and alert the driver to take back control. 
Yet, in reality, we would still greatly prefer a system that can make a safe decision in case the driver is inattentive\footnote{Driver inattention is a serious issue for semi-autonomous vehicles: studies have shown that the lack of active involvement correlates with both driver fatigue and tardy reactions to take-over requests~\citep{vogelpohl2019asleep}.} at the time and cannot take back control.
In such a case, we require evaluating what a system would have done in situations where it decided to abstain.
\end{example}

\begin{example}[Comparing ML radiologist assistants]\label{ex:triage}
Suppose that a hospital is evaluating third-party radiology application programming interfaces (API) that can assist with its diagnosis system. 
Each API will either give a prediction or abstain from making one; if it abstains, then a human radiologist will examine the input~\citep{raghu2019algorithmic}. 
The hospital is wary that there are inputs for which the professional would also abstain or have cognitive biases against~\citep{busby2018bias,madras2018predict}.
Thus, it would need to occasionally rely on the classifier's predictions even on examples that it chose to abstain.
If these ``hidden predictions'' are not readily available from the third-party providers (e.g., require extra costs), how can the hospital evaluate and compare their services?
\end{example}

\begin{example}[Evaluating an abstaining classifier's internal biases]\label{ex:bias}
Suppose that an independent agency is auditing an ML-based recidivism prediction system\footnote{Algorithmic approaches to recidivism prediction, such as COMPAS, are both popular and controversial.} that has been deployed for a certain amount of time.
Given the high stakes of misclassification, the system is equipped with a learned abstention mechanism and decides to occasionally abstain from making a prediction, in which case the rejected cases are examined by human judges.
The auditing agency is interested in checking whether the ML classifier possesses internal biases against certain demographic groups, and in particular, it wants to estimate the classifier's accuracy on each demographic group \emph{had it not abstained on any input}.  
While the agency has access to the system's past predictions and abstentions, it does not have access to the underlying predictive model or its abstention mechanism.
In other words, the agency requires a black-box evaluation method that estimates the counterfactual score of this system.
\end{example}

\subsection{An equivalent formulation in the potential outcomes framework}\label{sec:potential_outcomes}

There are other equivalent ways to formulate our setup (Section~\ref{sec:identification_cfscore}) using variants of the potential outcomes framework.
First, we can define a (potentially observed) prediction $f(X; R)$, which equals $f(X)$ if $R=0$ and $*$ if $R=1$, where the symbol $*$ indicates an abstention (the same notation is used in \citet{rubin1976inference}'s missing data framework). 
The score $S$ is then $\sfs(f(X), Y)$ if $R=0$ and $*$ otherwise. 

Alternatively, we can explicitly invoke \citet{rubin1974estimating}'s potential outcomes framework to write $S(0) \leftarrow \sfs(f(X), Y)$ and $S(1) \leftarrow *$, where $S(r)$ refers to the score of the abstaining classifier when $R=r$ for each $r \in \{0, 1\}$. 
We do not use this notation in our main paper because $S(1)$ is not meaningful in our case.

\subsection{Comparison with Condessa et al.'s score}\label{sec:condessa}

To better understand the counterfactual score $\psi = \mathbb{E}[S]$, we can contrast it with \citet{condessa2017performance}'s notion of the `classification quality score' $\theta$.
Assuming that $S \in [0,1],$ their \emph{classification quality score} $\theta$ can be defined as follows:
\begin{equation}\label{eqn:condessa_score}
\theta:= \ex{S\mid R = 0}\Psymb(R = 0) + \ex{1-S\mid R = 1}\Psymb(R = 1).
\end{equation}
In contrast, note that the counterfactual score~\eqref{eqn:cf_score} is decomposed into
\begin{equation}\label{eqn:cf_score_decomp}
\psi = \ex{S \mid R= 0} \Psymb(R = 0) + \ex{S\mid R = 1}\Psymb(R=1).
\end{equation}
Thus, our target quantity $\psi$ is large if the classifier is good on all inputs (abstentions or not), while $\theta$ is large if the classifier is good on points it predicts on but poor on points it abstains on.

\newlyadded{
\paragraph{A concrete example}
To further elucidate the difference in what each score measures, we include a hypothetical example.
Consider comparing two abstaining classifiers $\sfA$ and $\sfB$ based on their accuracy score over $n=100$ data points, whose inputs $(X_1, X_2)$ are sampled uniformly on $[-1, 1] \times [-1, 1]$. 

Suppose that the base classifier for $\sfA$ achieves a $S^\sfA=1.0$ accuracy when $X_1 < 0$ (the ``left half'') but only a $S^\sfA = 0.8$ when $X_1 \geq 0$ (the ``right half''). 
So, it decides to abstain at an 80\% rate on the right half ($\pi^\sfA(x_1, x_2) = 0.8$ if $x_1 < 0$), while it does not abstain at all on the left half ($\pi^\sfA(x_1, x_2) = 0$ if $x_1 \geq 0$). 
Assuming for the sake of simplicity that $n/2=50$ points are placed in each half, the classifier makes 50 (out of 50) correct predictions on the left half, while on the right half, it makes 8 (out of 10) correct predictions and 40 abstentions, for which it would have been correct 80\% of the time. 
This classifier's overall selective accuracy is thus $(50+8)/(50+10) = 58/60 \approx 0.97$, while its coverage is $(50+10)/100 = 0.6$.

Plugging in the counts and probabilities to~\eqref{eqn:cf_score_decomp}, we can calculate the empirical\footnote{This is only `empirical' up to the sampling of $n$ data points; we do not need to estimate the counterfactuals in this hypothetical example because they are already known.} counterfactual score of classifier $\sfA$:
\begin{equation}
    \hat\psi^\sfA = \frac{50 + 8}{50 + 10} \cdot \frac{1.0 + 0.2}{2} + \frac{0 + 32}{0 + 40} \cdot \frac{0 + 0.8}{2} = \frac{58}{60} \cdot 0.6 + \frac{32}{40} \cdot 0.4 = 0.9.
\end{equation}
(A simpler calculation would be to use~\eqref{eqn:cf_score} directly, but we use the equivalent decomposition~\eqref{eqn:cf_score_decomp} here to contrast with~\eqref{eqn:condessa_score}.)
The empirical classification quality score of classifier $\sfA$ is
\begin{equation}
    \hat\theta^\sfA = \frac{58}{60} \cdot 0.6 + \inparen{1 - \frac{32}{40}} \cdot 0.4 = 0.66.
\end{equation}
Comparing the two scores, $\hat\psi^\sfA$ is much larger than $\hat\theta^\sfA$ because the classifier chose to abstain on 40\% of the data for which it would have gotten a $0.8$ accuracy.

Next, suppose that classifier $\sfB$ is the same as classifier $\sfA$, except that it achieves a meager $0.6$ accuracy on the right half. 
It also uses the same abstention mechanism as $\sfA$.
Then, analogous calculations show that classifier $\sfB$’s counterfactual score $\hat\psi^\sfB$ would be $56/60 \cdot 0.6 + 24/40 \cdot 0.4 = 0.8$, lower than $\hat\psi^\sfA$, whereas the classification quality score $\hat\theta^\sfB$ would be $56/60 \cdot 0.6 + (1-24/40) \cdot 0.4 = 0.72$, higher than $\hat\theta^\sfA$.
Thus, comparisons based on the two scores would lead to opposite conclusions.
Note that the classification quality score rewards $\sfB$ for hiding its low-accuracy predictions, even though $\sfA$ uses the same abstention mechanism and has a better accuracy overall on the right half.

The choice between the two scores should ultimately be determined by the use case, although we focus on the counterfactual score in the main paper, motivated by the various cases we described earlier (Example~\ref{ex:api} and the additional examples in Appendix~\ref{sec:examples}).
In the example above, if the evaluator needs to later access the classifier’s hidden predictions on the right half, then they can use the counterfactual score and choose $\sfA$, which has a higher accuracy in its hidden predictions than $\sfB$.
}

\paragraph{Estimation} As mentioned in Section~\ref{sec:introduction}, \citet{condessa2017performance} focus on the ``white-box'' setup where the hidden predictions are known to the evaluator, and it is not obvious how to estimate their score in the black-box setup.
However, much like the counterfactual score $\psi$, the challenge of estimating $\theta$ is driven entirely by the $\ex{S|R = 1}$ term, as the remaining terms are directly observed. 
As the decompositions~\eqref{eqn:condessa_score} and~\eqref{eqn:cf_score_decomp} show, estimates of $\psi$ (from the DR CI in Section~\ref{sec:estimation_cfscore}) also yield estimates of $\theta$, since $\theta + \psi$ is an observable quantity that can be straightforwardly estimated.
Subtracting an estimate of $\psi$ from the sum gives an estimate of $\theta$.


\subsection{The plug-in and inverse propensity weighting estimators}\label{sec:baselines}

The uniqueness of efficient influence functions tells us that the DR estimator outperforms two intuitive yet suboptimal estimators in an asymptotic and locally minimax sense. 
The first is the \emph{plug-in estimator}, which is derived directly from the identified target $\psi = \mathbb{E}[\mu_0(X)]$ in Proposition~\ref{ppn:identification_cfscore}:
\begin{equation}\label{eqn:plugin}
\hat\psi_\plugin = \frac{1}{n}\sum_{i=1}^n \hat\mu_0(X_i),
\end{equation}
where $\hat\mu_0$ is any estimate of the regression function $\mu_0(x) = \condex{S}{R=0, X=x}$. 
The quality of this simple estimator directly depends on the estimation quality of $\hat\mu_0$ for $\mu_0$, and in a nonparametric setting, the estimator can suffer from the statistical curse of dimensionality.
Another point of concern is that it makes no use of the missingness patterns.

The second is \emph{inverse probability weighting (IPW)} estimator~\citep{horvitz1952generalization,rosenbaum1995design}: 
\begin{equation}\label{eqn:ipw}
\hat\psi_\ipw = \frac{1}{n}\sum_{i=1}^n \frac{(1-R_i)}{1-\hat\pi(X_i)} S_i,
\end{equation}
where $\hat\pi$ is an estimate of the abstention mechanism $\pi(x) = \condprob{R=1}{X=x}$.
If $\hat\pi$ consistently estimates $\pi$, the IPW estimator is unbiased; yet, it has the opposite problem to the plug-in estimator as it does not model the conditional score $\mu_0$ at all.


\subsection{Positivity and policy}\label{sec:policy}

Our identification results in Section~\ref{sec:identification_cfscore} impose a requirement of positivity (Assumption~\ref{assn:positivity_cfscore}) on the abstaining classifier $(f,\pi)$, i.e., a demand that for some $\epsilon > 0,$ the essential supremum of $\pi(x)$ is smaller than $1-\epsilon$. This requirement is necessary: intuitively, if no feedback about the behaviour of $f$ is available in a region, it is impossible (without further strong assumptions about the global structure of $f$) to determine the behaviour of the score in this region. Operationally, this is seen quite directly in the validity of the confidence intervals inferred from data (Figure~\ref{fig:positivity}). 
Of course, the parameter $\epsilon$ also plays a quantitative role: the higher the $\epsilon,$ the better the validity and widths of our CIs. In other words, our ability to identify decays gracefully with $\epsilon$, with complete inability if $\pi(x) = 1$ in a region of large mass.

While necessary, this positivity requirement is at odds with the practical deployment of client-facing abstaining classifiers. Indeed, there are two major reasons to implement an abstaining mechanism in such scenarios. In a positive sense, abstentions signal that the use of the underlying classifier $f$ is inappropriate in a particular domain. However, in a negative sense, abstentions can also be employed in order to artificially limit a vendor's liability when their predictions (and the actions driven by the same) are incorrect. A pertinent example is the recent investigation of the Tesla autopilot by the \citet{Nhtsa_tesla} which found that in 16 incidents, the autopilot would deactivate and hand-off control to the driver at the very last seconds before a crash, thus artificially inflating the safety metrics of the system. 

Part of the impetus behind studying a metric such as the counterfactual score is precisely to identify such behaviours before unsafe incidents bring them to light. Nevertheless, if vendors can stymie this investigation simply by ensuring that abstention is accompanied by a very high $\pi(x),$ then the method is not particularly useful. 

This technical impasse begs for a policy-level treatment: through regulatory action, the executive may ensure that vendors supply evaluators (whether government agencies or independent reviewers) with abstaining classifiers that reveal the counterfactual decision of $f$ at least an $\epsilon$-fraction of the times when the decision is to abstain, where $\epsilon$ is set by mutual agreement of the stakeholders. Note that it is not enough to just supply evaluators with the predictions of $f$ (although this would solve our particular problem formulation), since it is important to understand its behaviour in the context of when the abstaining classifier actually tends to reject points (i.e., it is equally important for evaluators and users to understand $\ex{S\mid R= 1}$, which of course is estimable under our setup).


\newlyadded{
\subsection{Extension to learning-to-defer settings}\label{sec:learning_to_defer}
In the learning-to-defer setting involving an expert~\citep{madras2018predict}, the counterfactual score would refer to the expected score of the overall system had the classifier not deferred at all.
The counterfactual score is thus an evaluation metric primarily for the classifier, and it is independent of the expert’s predictions, even when the classifier is adaptive to the expert’s tendencies.

On the other hand, in the case where the goal is to assess the \emph{joint} performance of the algorithm and the expert, then it may be useful to estimate a variation of the classification quality score~\eqref{eqn:condessa_score} defined in Appendix~\ref{sec:condessa}. 
If we denote the expert’s score as $E$, then equation~\eqref{eqn:condessa_score} can further be generalized to
\begin{equation}
\theta^E := \mathbb{E}[S \mid R=0] \mathbb{P}(R=0) + \mathbb{E}[E - S \mid R=1] \mathbb{P}(R=1).    
\end{equation}
For each rejection ($R=1$), the score would assess the system by the difference in the quality of expert prediction and the model prediction ($E-S$); in the black-box evaluation case, the model prediction score in the case of deferral ($S$ given $R=1$) is a counterfactual.

Finally, the estimation approach from Appendix~\ref{sec:condessa} still applies to $\theta^E$.
If the expert is an oracle ($E=1$), then $\theta^E$ coincides with the classification quality score $\theta$~\eqref{eqn:condessa_score}. 
Even if the expert's predictions are random, $\mathbb{E}[E \mid R=1]$ is an observable quantity and $\theta^E$ can be re-written as $\theta + \mathbb{E}[E-1 \mid R=1]\Psymb(R=1)$, so $\theta^E$ can be estimated within our counterfactual framework.
}


\section{Proofs}\label{sec:proofs}

\subsection{Proof of Proposition~\ref{ppn:eval_indep_mar}}\label{sec:proof_eval_indep_mar}
Since $(X, Y)$ is independent of the training data $\calD_{\mathrm{train}}$ for $(f, \pi)$, and because $\xi$ is an independent source of randomness, we can treat the functions $f$ and $\pi$ as fixed.
Then, by definition, $S = \sfs(f(X), Y)$ is a deterministic function of $(X, Y)$ and $R = \sfr(\pi(X), \xi)$ is a deterministic function of $X$ and $\xi$.
This means that the condition $S \indep R \mid X$ is equivalent to saying that $Y \indep \xi \mid X$.
Given that $\xi$ is independent of $(X, Y)$, the latter condition follows.

\subsection{Proof of Proposition~\ref{ppn:identification_cfscore}}\label{sec:proof_identification_cfscore}

Positivity (Assumption~\ref{assn:positivity_cfscore}) ensures that the conditional expectation $\mu_0(X) = \condex{S}{R=0, X}$ is well-defined.
Then,
\begin{align}\label{eqn:identification_proof}
\ex{\mu_0(X)} = \ex{\condex{S}{R=0, X}} \overset{\mathrm{(MAR)}}{=} \ex{\condex{S}{X}} = \ex{S} = \psi,
\end{align}
where the second inequality follows from the MAR condition (Assumption~\ref{assn:mar_cfscore}), i.e., $S \indep R \mid X$.

\subsection{Proof Sketch of Theorem~\ref{thm:evaluation_cfscore}}\label{sec:proof_evaluation_cfscore}

We follow the relevant notations and derivations from \citet{kennedy2022semiparametric}.
Denote $\Psymb\incurly{f} = \Ex{\Psymb}{f(Z)}$ and $\Psymb_n\incurly{f} = n^{-1}\sum_{i=1}^n f(Z_i)$ where $Z_i \overset{iid}{\sim} \Psymb$.
We use the \emph{centered} influence function for $\psi(\Psymb) = \Ex{\Psymb}{\mu_0(X)}$ (upon identification), defined as follows:
\begin{equation}\label{eqn:influence_centered}
    \influence_\Psymb(x, r, s) := \insquare{\frac{1-r}{1-\pi(x)}\inparen{s - \mu_0(x)} + \mu_0(x)} - \psi(\Psymb).
\end{equation}
Here, $\influence_\Psymb$ depends on $\Psymb$, which determines $\pi$ and $\mu_0$.
Analogously, we let $\hat\Psymb$ denote the distribution of abstentions and score outcomes involving estimators $\hat\pi$ and $\hat\mu_0$ (in place of $\pi$ and $\mu_0$), and let $\influence_{\hat\Psymb}$ and $\psi(\hat\Psymb)$ denote the corresponding influence function and target functional, respectively, defined using $\hat\pi$ and $\hat\mu_0$.
Also, note that an uncentered version is shown in the main text for ease of explanation; the resulting variance does not change due to this centering.
Using these definitions, we proceed with the proof in two steps.

\paragraph{Step 1: Showing that $\influence$~\eqref{eqn:influence_centered} is the efficient influence function for $\psi$}
To show that $\influence$ is indeed the unique efficient influence function for $\psi$, we show that $\Psymb\incurly{\influence_\Psymb} = 0$ and that its bias term is second-order. 
The uniqueness and asymptotic efficiency of this EIF in a nonparametric setting, in general, is well-known (e.g., \citet{vandervaart2002semiparametric}).
First, observe that
\begin{align}
    \Psymb\incurly{\influence_\Psymb} 
    &= \Ex{\Psymb}{\frac{1-R}{1-\pi(X)}\inparen{S - \mu_0(X)} + \mu_0(X)} - \psi(\Psymb) \\
    &= \Ex{\Psymb}{ \frac{\ex{(1-R)(S-\mu_0(X)) \mid X}}{1-\pi(X)} } \\
    &\overset{\mathrm{(a)}}{=} 0,
\end{align}
where $\mathrm{(a)}$ follows from the fact that
\begin{align}
    \ex{(1-R)S\mid X} = \pi(X) \cdot 0 + (1-\pi(X)) \ex{S\mid R=0,X} = (1-\pi(X))\mu_0(X).
\end{align} 

Furthermore, for any distributions $\hat\Psymb$ and $\Psymb$, the bias term is given by
\begin{align}\label{eqn:bias_evaluation}
    R_2(\hat\Psymb, \Psymb) &= \psi(\hat\Psymb) - \psi(\Psymb) + \Psymb\incurly{\influence_{\hat\Psymb}} \\
    &= \psi(\hat\Psymb) - \psi(\Psymb) + \Ex{\Psymb}{\frac{1-R}{1-\hat\pi(X)}\inparen{S - \hat\mu_0(X)} + \hat\mu_0(X)} - \psi(\hat\Psymb) \\
    &= \Ex{\Psymb}{\frac{1-R}{1-\hat\pi(X)}\inparen{S - \hat\mu_0(X)} + \hat\mu_0(X) - \mu_0(X)} \\
    &\overset{\mathrm{(IE,a)}}{=} \Ex{\Psymb}{\frac{1-\pi(X)}{1-\hat\pi(X)} \inparen{\mu_0(X) - \hat\mu_0(X)} - \inparen{\mu_0(X) - \hat\mu_0(X)}} \\
    &= \Ex{\Psymb}{\frac{\inparen{\hat\pi(X)-\pi(X)}\inparen{\mu_0(X)-\hat\mu_0(X)}}{1-\hat\pi(X)}} \\
    &\leq \frac{1}{\epsilon} \cdot \norm{\hat\pi - \pi}_{L_2(\Psymb)} \norm{\hat\mu_0 - \mu_0}_{L_2(\Psymb)}. \label{eqn:if_bias}
\end{align}
This is a second-order product term in the difference of $\hat\Psymb$ and $\Psymb$, showing that $\influence$ is an influence function for $\Psymb$. 

\paragraph{Step 2: Showing the asymptotic normality of $\sqrt{n}(\hat\psi_{\dr} - \psi)$} 
To derive the explicit form of the limiting distribution, denote $\hat\influence = \influence_{\hat\Psymb}$, and observe that the DR estimator is a ``one-step'' bias-corrected estimator~\citep{bickel1975onestep}, given by $\hat\psi_\dr = \Psymb_n\{\hat\influence\} + \psi(\hat\Psymb)$. 
Then, we have the following three-term decomposition:
\begin{align}
    \hat\psi_\dr - \psi &= \Psymb_n \incurly{\hat\influence} + \psi(\hat\Psymb) - \psi(\Psymb) \\
    &= \inparen{\Psymb_n - \Psymb}\incurly{\hat\influence} + R_2(\hat\Psymb, \Psymb) \\
    &= \inparen{\Psymb_n - \Psymb}\incurly{\influence} + \inparen{\Psymb_n - \Psymb}\incurly{\hat\influence - \influence} + R_2(\hat\Psymb, \Psymb).
\end{align}
The first term, which is a sample average term, has the desired limiting distribution by the central limit theorem:
\begin{equation}
   \sqrt{n} \cdot \inparen{\Psymb_n - \Psymb}\incurly{\influence} = \frac{1}{\sqrt{n}} \sum_{i=1}^n \insquare{\influence(Z_i) - \Ex{\Psymb}{\influence(Z)}} \rightsquigarrow \calN\inparen{0, \Var{\Psymb}{\influence}}.
\end{equation}
Then, by Slutsky's theorem, it suffices to show that the other two terms are of order $o_\Psymb(1/\sqrt{n})$.
The third term, $R_2(\hat\Psymb, \Psymb)$, is precisely the second-order bias term we derived in~\eqref{eqn:if_bias}, and it is $o_\Psymb(1/\sqrt{n})$ by the DR assumption~\eqref{eqn:dr_assn}.

The second term, called the empirical process term, can be shown to be of order $o_\Psymb(1/\sqrt{n})$ when using cross-fitting to estimate $\hat\Psymb$.
Specifically, the sample splitting procedure guarantees that $\hat\Psymb \indep \Psymb_n$ (where $\Psymb_n$ now refers to the held-out fold in each step of cross-fitting), which is enough to show that
\begin{equation}
\inparen{\Psymb_n - \Psymb}\incurly{\hat\influence - \influence} = O_\Psymb\inparen{\frac{\lVert{\hat\influence - \influence}\rVert_{L^2(\Psymb)}}{\sqrt{n}}}.
\end{equation}
Since $\lVert{\hat\influence - \influence}\rVert_{L^2(\Psymb)} = o_\Psymb(1)$ by assumption, the term itself is of order $o_\Psymb(1/\sqrt{n})$ as desired.
The loss of sample efficiency due to a single sample splitting can be recovered by the cross-fitting procedure.
See, e.g., Lemma 1 and Proposition 1 of~\citet{kennedy2022semiparametric} for details.

\subsection{Proof of Theorem~\ref{thm:comparison_cfscore}}\label{sec:proof_comparison_cfscore}

Given that $\influence_\Psymb^{\sfA\sfB} = \influence_\Psymb^{\sfA} - \influence_\Psymb^{\sfB}$, it is immediate that it is an influence function for $\Delta^{\sfA\sfB} = \psi^\sfA - \psi^\sfB$ because $\Psymb \{\influence_\Psymb^{\sfA\sfB}\} = \Psymb\{\influence_\Psymb^{\sfA}\} - \Psymb\{\influence_\Psymb^{\sfB}\} = 0$ and
\begin{equation}\label{eqn:bias_comparison}
R_2(\hat\Psymb, \Psymb) \leq \frac{1}{\epsilon} \cdot \inparen{ \norm{\hat\pi_\sfA - \pi_\sfA}_{L_2(\Psymb)}\norm{\hat\mu_{0, \sfA} - \mu_{0, \sfA}}_{L_2(\Psymb)} + \norm{\hat\pi_\sfB - \pi_\sfB}_{L_2(\Psymb)}\norm{\hat\mu_{0, \sfB} - \mu_{0, \sfB}}_{L_2(\Psymb)} }.
\end{equation}
The limiting distribution can also be derived analogously, where the upper bound in~\eqref{eqn:bias_comparison} reveals the additive form of the DR assumption~\eqref{eqn:dr_assn_comparison}.


\section{Illustration of the MAR condition via causal graphs}\label{sec:mar_graph}

\begin{figure}[h!]
    \centering
    \begin{subfigure}[t]{0.45\textwidth}
        \centering
        \begin{tikzpicture}
        \node[obs] (X) {$X$};
        \node[obs, right=1.2cm of X] (Y) {$Y$};
        \node[obs, below=of X] (R) {$R$};
        \node[det, below=of Y] (S) {$S$};
        \edge {X} {R, Y};
        \edge {X, Y} {S};
        \end{tikzpicture}
        \caption{Simple DAG representation of our setup, assuming $X \to Y$.}
        \label{subfig:mar_graph_simple}
    \end{subfigure}
    \hspace{2em}
    \begin{subfigure}[t]{0.45\textwidth}
        \centering
        \begin{tikzpicture}
        \node[obs] (X) {$X$};
        \node[obs, right=1.2cm of X] (Y) {$Y$};
        \node[obs, below=of X] (R) {$R$};
        \node[det, below=of Y] (S) {$S$};
        \node[latent, left=1.2cm of R] (xi) {$\xi$};
        \node[latent, above=0.8cm of X, xshift=0.9cm] (U) {$U$};
        \path (X) edge [dashed] (Y);
        \edge {X, xi} {R};
        \edge {U} {X, Y};
        \edge {X, Y} {S};
        \end{tikzpicture}
        \caption{A more general graph that allows arbitrary relationships between $X$ and $Y$ as well as the classifier's internal randomness/bias ($\xi$).}
        \label{subfig:mar_graph_confounder}
    \end{subfigure}
    \caption{Two graphical representations of the random variables involved in our evaluation framework from Section~\ref{sec:cfscore}, assuming that the true label $Y$ is independent of the abstaining classifier $(f, \pi)$.
    Shaded variables are observed by the evaluator; the score $S = \sfs(f(X), Y)$ is \emph{partially} observed by the evaluator (depicted as a diamond node).
    In plot~\ref{subfig:mar_graph_simple}, assuming $X \to Y$, the simple DAG illustrates that $S$ and $R$ are $d$-separated given $X$.
    In plot~\ref{subfig:mar_graph_confounder}, we further allow arbitrary relationships between $X$ and $Y$, including $X \to Y$, $Y \to X$, and $U \to (X, Y)$ for some unobserved confounder $U$.
    The classifier's decision to abstain, $R$, is also allowed to additionally depend on some internal randomness and bias $\xi$ that is independent of the evaluation data.
    Accounting for these generalizations, $S$ and $R$ are still $d$-separated given $X$, irrespective of the causal direction (if any) between $X$ and $Y$.
    }
    \label{fig:mar_graph}
\end{figure}

Intuitively, the MAR condition is satisfied as long as the evaluation label is unknown to either classifier, simply because the classifier cannot access the actual score $S = \sfs(f(X),Y)$, which is a function of the true label $Y$, in making its abstention decision. This already implies $P(R=1 | S, X) = P(R=1 | X)$.
We can further elucidate how the causal relationships between the random variables in our setup, and highlight how the MAR condition is generally satisfied, via graphical representations of the evaluation setup.
The comparison case is an analogous extension to two abstaining classifiers.

Assuming that the abstaining classifier $(f, \pi)$ does not depend on the evaluation output label $Y$, we (the evaluator) can treat both functions as fixed given the input $X$.
We can then illustrate the MAR condition via two causal graphs.
First, suppose $X \to Y$ (for the sake of simplification).
Then, we have the relationships $X \to Y$, $X \to R$ (Bernoulli with probability $\pi(X)$), and $(X, Y) \to S$ (deterministic via $f$ and $\sfs$).
In the resulting graph, shown in Figure~\ref{subfig:mar_graph_simple}, the variables $S$ and $R$ are $d$-separated~\citep{pearl2000models} given $X$, i.e., $S \indep R \mid X$.
Note that $S$ is partially observed and thus drawn as a diamond node, but it does not affect the conditional independence relationship. 
An alternative representation is possible via missingness graphs~\citep{mohan2013graphical}, which would give us the same conclusion.

Next, we can remove the assumption on the relationship $X \to Y$, and allow any possible relationship between $X$ and $Y$: $X \to Y$ (causal), $Y \to X$ (anticausal), or $U \to (X, Y)$, where $U$ is an unobserved confounder to the prediction task. 
This is depicted as a dashed line between $X$ and $Y$, along with a possible presence of $U$, in Figure~\ref{subfig:mar_graph_confounder}.
We can further allow the abstaining classifier to utilize some internal randomness or bias $\xi$, which is independent of the randomness in evaluation data, for its decision to abstain $R$.
In the resulting graph, shown in Figure~\ref{subfig:mar_graph_confounder}, none of the generalizations change the fact that $S$ and $R$ are $d$-separated given $X$, i.e., the MAR condition is satisfied.

Finally, as mentioned in the main text, the MAR condition can be violated when the evaluation data is not independent of the training data.
For example, if the true label $Y$ is used by the abstaining classifier during its training to inform its abstention decision, then this would correspond to a graph in which there is an additional edge from $Y$ to $R$, as the abstention function $\pi$ now depends on $Y$.
Then, $S$ and $R$ are no longer $d$-separated because there is a now connecting path via $Y$ (common cause).


\section{Confidence sequences for anytime-valid counterfactual score estimation}\label{sec:asympcs}

The nonparametric efficiency result of Theorem~\ref{thm:evaluation_cfscore} yields an optimal inference procedure (either a hypothesis test or a confidence interval) for evaluating and comparing abstaining classifiers at a fixed sample size.
Here, we go one step further and utilize a \emph{confidence sequence (CS)}~\citep{darling1967confidence,howard2021timeuniform}, which is a sequence of confidence intervals whose validity holds uniformly over all sample sizes.
This \emph{time-uniform} property allows the evaluator to continuously monitor the result as more data is collected over time. 
The time-uniform property also implies \emph{anytime-validity}~\citep{johari2021always,grunwald2019safe}, which allows the evaluator to run the experiment without pre-specifying the size of the evaluation set and compute the CIs as more data is collected. 
This implies that anytime-valid methods avoid the issue of inflated miscoverage rates coming from ``data peeking.''
See~\citet{ramdas2022game} for an introduction.

Formally, for any $\alpha \in (0,1)$, a $(1-\alpha)$-level (non-asymptotic) CS $(C_t)_{t\geq 1}$ for a parameter $\theta \in \R$ is a sequence of confidence intervals (CI) such that
\begin{equation}\label{eqn:cs}
    \prob{\forall t \geq 1: \theta \in C_t} \geq 1 - \alpha.
\end{equation}
Importantly, a CS contrasts with a fixed-time CI, whose guarantee no longer remains valid at stopping times: a CI only satisfies $\prob{\theta \in C_t} \geq 1 - \alpha$ for a fixed sample size $t$.

Here, we describe how we can perform the proposed counterfactual comparison of abstaining classifiers using a variant of a CS that is asymptotic and readily applicable to causal estimands~\citep{waudbysmith2021doubly}.
An (two-sided) \emph{$(1-\alpha)$-asymptotic CS (AsympCS)} $(\tilde{C}_t)_{t\geq1}$ for a parameter $\theta \in \R$ is a sequence of intervals, $\tilde{C}_t = (\hat\theta_t \pm \tilde{B}_t)$, for which there exists a non-asymptotic CS $(C_t)_{t\geq 1}$ for $\theta$ of the form $C_t = (\hat\theta_t \pm B_t)$ that satisfies
\begin{equation}\label{eqn:asympcs_defn}
B_t / \tilde{B}_t \overset{\mathsf{a.s.}}{\longrightarrow} 1.
\end{equation}
The AsympCS has an \emph{approximation rate} of $r_t$ if $\tilde{B}_t - B_t = O(r_t)$ almost surely.

Intuitively, an AsympCS is an arbitrarily precise approximation of a non-asymptotic CS.
Because no known non-asymptotic CS exists for counterfactual quantities such as the ATE, AsympCS has been derived as an (only) viable alternative.
\citet{waudbysmith2021doubly} further leverage the (previously described) nonparametric efficiency theory and doubly robust estimation to derive an AsympCS for the ATE in randomized experiments and observational studies; we apply their theory to estimating the counterfactual scores and their differences.
The resulting AsympCS is asymptotically time-uniform and anytime-valid, and its width scales similarly, up to logarithmic factors, to a fixed-time CI derived directly from Theorem~\ref{thm:evaluation_cfscore}.

Now we describe our main theorem for anytime-valid and counterfactual evaluation of an abstaining classifier.
We consider evaluating the classifier on an i.i.d. test set that is continuously collected over time; let $n$ be the (data-dependent) sample size with which inference is performed.
As before, the comparison problem reduces to evaluating each abstaining classifier and taking their difference.
We suppose that the nuisance functions $\hat\pi$ and $\hat\mu_0$ are learned via cross-fitting, and these are used to compute the EIF estimate~\eqref{eqn:dr}.
Now we can formally state an asymptotic CS for $\psi = \mathbb{E}[S]$~\eqref{eqn:cf_score} that is anytime-valid and doubly robust.
In the below, the $o(\cdot)$ notation refers to almost sure convergence.
\begin{theorem}[Anytime-valid DR estimation of the counterfactual score]\label{thm:asympcs}
Suppose that $\hat\mu_0$ and $\hat\pi$ consistently estimates $\mu_0$ and $\pi$ in $L_2(\Psymb)$, respectively, at a product rate of $o(\sqrt{\log\log n / n})$:
\begin{equation}\label{eqn:assn_dr_asympcs}
\norm{\hat\mu_0 - \mu_0}_{L_2(\Psymb)} \norm{\hat\pi - \pi}_{L_2(\Psymb)} = o(\sqrt{\log\log n / n}).
\end{equation}
Also, suppose that $\lVert\hat\influence - \influence\rVert_{L_2(\Psymb)} = o(1)$ and that $\influence$ has at least four finite moments.

Then, under Assumption~\ref{assn:mar_cfscore} and~\ref{assn:positivity_cfscore}, for any choice of $\rho > 0$,
\begin{equation}\label{eqn:asympcs}
    \hat\psi_\dr \pm \sqrt{\hat{\mathsf{Var}}_n\inparen{\hat{\influence}}} \cdot \sqrt{ \frac{2n\rho^2 + 1}{n^2\rho^2} \log\inparen{ \frac{\sqrt{n\rho^2+1}}{\alpha}} }
\end{equation}
forms a $(1-\alpha)$-AsympCS for $\psi$ with an approximation rate of $\sqrt{\log\log n / n}$.
\end{theorem}
This result is an adaptation of Theorems 2.2 and 3.2 in~\citet{waudbysmith2021doubly} to our setup.
The assumptions on $\hat\pi$ and $\hat\mu_0$ are analogous to the double robustness assumptions~\eqref{eqn:dr_assn} in Theorem~\ref{thm:evaluation_cfscore}, as they require the same product rate up to logarithmic factors.
Here, $\rho$ is a free parameter that can be chosen to optimize the CS width (see Appendix C.3 of \citet{waudbysmith2021doubly} for details).

Compared to the fixed-size CI of~\eqref{thm:evaluation_cfscore}, whose width shrinks at a $O(1/\sqrt{n})$ rate, the width of the AsympCS in~\eqref{eqn:asympcs} shrinks at a $O(\sqrt{\log n / n})$ rate. 
This means that, in terms of the CI width, the extra cost of ensuring anytime-validity is logarithmic in $n$. 
In practice, the AsympCS may be wider than the CI from Theorem~\ref{thm:evaluation_cfscore}; nevertheless, the AsympCS may be preferred in scenarios where the evaluation/comparison is performed on continuously collected data.
Another potential benefit of the AsympCS is the extension to settings with sequential and time-varying evaluation tasks (e.g., involving time-series forecasters that abstain). We leave the formalization of the time-varying setup as future work.

Finally, to apply Theorem~\ref{thm:asympcs} to a comparison setting, we can construct two $(1-\alpha/2)$-AsympCSs, $C_n^\sfA = (L_n^\sfA, U_n^\sfA)$ and $C_n^\sfB = (L_n^\sfB, U_n^\sfB)$ for $\psi^\sfA$ and $\sfB$ respectively, and then combine them into one $(1-\alpha)$-AsympCS for $\Delta^{\sfA\sfB} = \psi^\sfA - \psi^\sfB$ via $C_n^{\sfA\sfB} = (L_n^\sfA - U_n^\sfB, U_n^\sfA - L_n^\sfB)$.


\section{Additional experiments and details}\label{sec:additional_experiments}

\subsection{Details on the simulated data and abstaining classifiers}\label{sec:binary_mar_setup}

The evaluation set is generated as follows: $(X_{0i}, X_{1i}) \sim \mathsf{Unif}[0,1]$, $E_i \sim \mathsf{Ber}(0.15)$, and $Y_i = \indicator{X_{0i} + X_{1i} \geq 1}$ if $E_i=0$ and $Y_i = \indicator{X_{0i} + X_{1i} < 1}$ otherwise (label noise). 
Classifier $\sfA$ uses a logistic regression model with the optimal linear decision boundary, i.e., $f^\sfA(x_0, x_1) = \sigma(x_0 + x_1 - 1)$, where $\sigma(u) = 1/(1+\exp(-u))$, achieving an accuracy of $0.85$ by design.
Classifier $\sfB$, on the other hand, has a (suboptimal) curved boundary: $f^\sfB(x_0, x_1) = 0 \vee (\frac{1}{2}(x_0^2 + x_1^2) + \frac{1}{10}) \wedge 1$. 
Classifier $\sfA$ is thus ``oracle'' logistic regression model with the same decision boundary, achieving an empirical score of $0.86$ before abstentions; classifier $\sfB$ is a biased model that achieves an empirical score of $0.74$ before abstentions.

For both classifiers, $\epsilon = 0.2$ determines the coefficient for positivity, and they are designed to abstain more frequently near their decision boundaries.
For classifier $\sfA$, $\pi^\sfA(x) = 1-\epsilon$ if the distance from $x$ to its boundary is less than $\delta$, and $\pi^\sfA(x) = \epsilon$ otherwise; for classifier $\sfB$, we use $0.8\delta$ as the threshold, resulting in less abstentions than $\sfA$.
In some sense, this is a setting where $\epsilon$-positivity is ``minimally'' satisfied because the abstention rate is always either $\epsilon$ or $1-\epsilon$, and not in between, in all regions of the input space.
If, say, the abstention rate was 0.5 in most parts but $\epsilon$ in a small region, the positivity level would still be $\epsilon$ but the estimation would in general be easier.
Thus, this example can be viewed as a more challenging case than a standard causal inference setup with small regions of $\epsilon$-positivity.

\begin{figure}[t]
    \centering
    \includegraphics[height=0.25\textwidth]{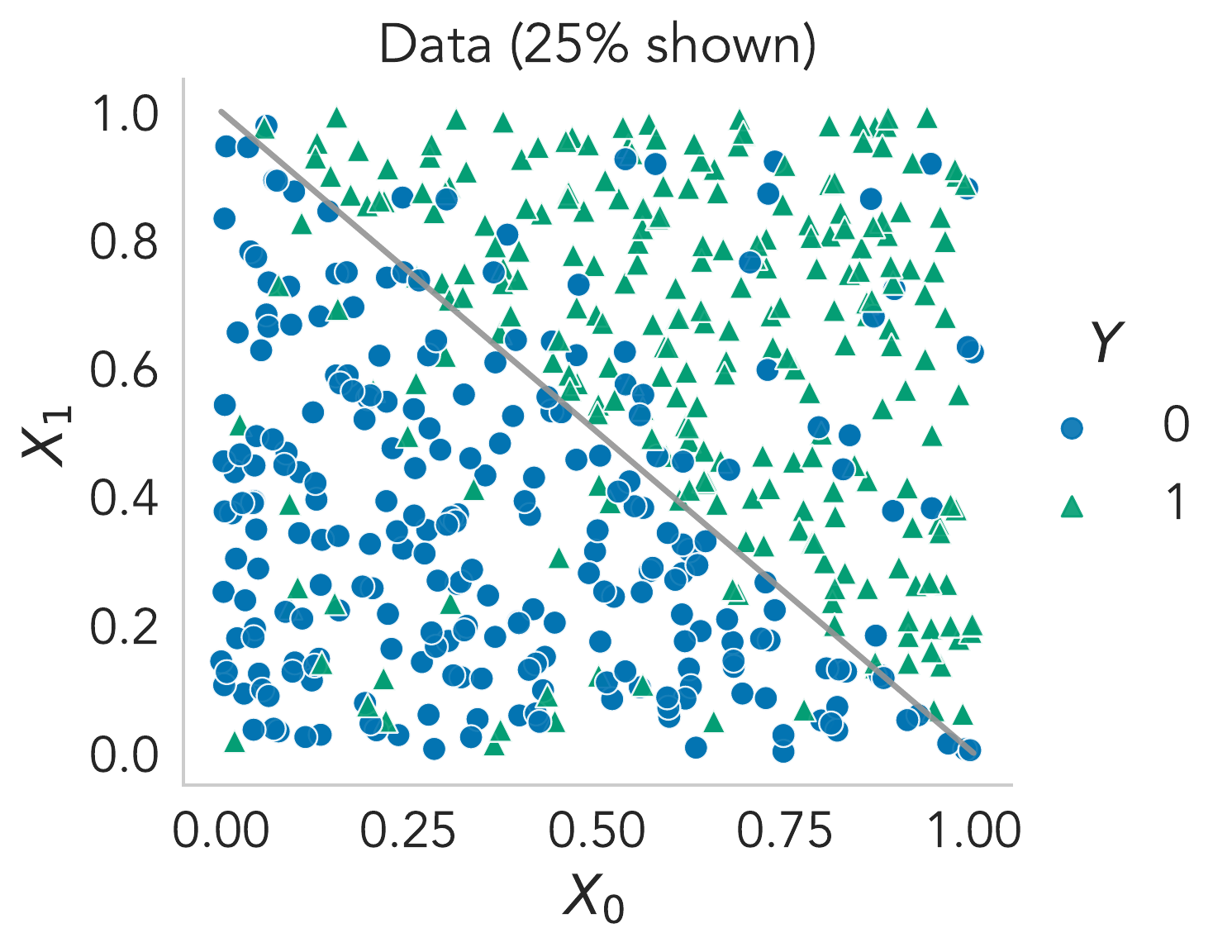}
    \includegraphics[height=0.25\textwidth]{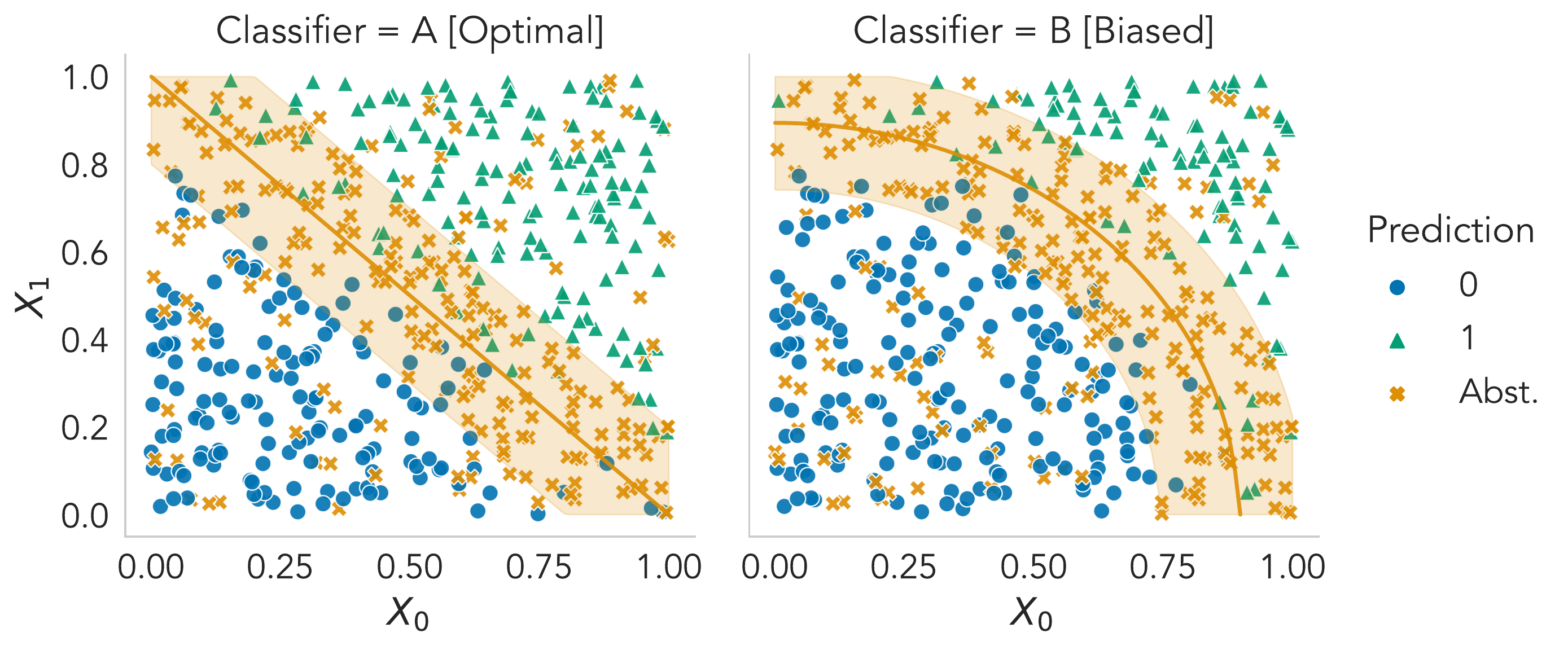}
    \caption{A simulated example where we compare two hypothetical abstaining classifiers. 
    The left plot shows a binary classification dataset (25\% shown) in which the true decision boundary is linear.
    The two plots on the right show both the predictions (blue circles for 0; green triangles for 1) and the abstentions (orange x's) of two classifiers: $\sfA$, which has the optimal linear boundary, and $\sfB$, which has the biased nonlinear boundary. 
    Both classifiers abstain w.p. $1-\epsilon$ in the shaded (orange) region near the decision boundary and w.p. $\epsilon$ outside the region.
    For both classifiers, $\epsilon$ is set to $0.2$ (positivity is satisfied).
    Because the abstention mechanism of either classifier is determined by the input, it is not uniformly spread out across the input domain (MAR).
    As a result, the difference in \emph{selective scores}, i.e., $\mathbb{E}[S^\sfA \mid R^\sfA=0] - \mathbb{E}[S^\sfB \mid R^\sfB=0] \approx 0.044$, is substantially smaller than the difference in the \emph{counterfactual scores}, i.e., $\Delta^{\sfA\sfB} = \mathbb{E}[S^\sfA - S^\sfB] \approx 0.116$.
    Our 95\% DR CI for $\Delta^{\sfA\sfB}$ the yields $(0.077, 0.145)$, using $n=2,000$.}
    \label{fig:binary_mar}
\end{figure}

Figure~\ref{fig:binary_mar} shows both the predictions (blue circles: 0, green triangles: 1) and the abstention decisions (orange x's: predictions) for each classifier.
Each classifier has a high chance of abstaining near its boundary (shaded orange region) and a low chance otherwise, meaning that abstentions are \emph{not} spread out uniformly (MAR but not MCAR).
In particular, classifier $\sfB$ hides many of its misclassifications as abstentions, leading to its high selective score ($\mathsf{Sel}^\sfB = 0.81$) relative to its counterfactual score ($\psi^\sfB = 0.74$).

The nuisance functions $\hat\pi$ and $\hat\mu_0$ for each classifier $\sfA$ and $\sfB$ are learned via 2-fold cross-fitting.
In each case, we cap extreme propensity predictions by $\hat\pi^\sfA$ and $\hat\pi^\sfB$ are capped at $1-\epsilon$.

On a 128-core CPU machine, using parallel processing, the entire compute time it took to produce Table~\ref{tbl:miscoverage_bias} was approximately 5 minutes.

\subsection{Power analysis}\label{sec:power}

To examine the efficiency of the DR estimator, we now analyze the power of the statistical test for $H_0: \Delta^{\sfA\sfB} = 0$ vs. $H_1: \Delta^{\sfA\sfB} \neq 0$ by inverting the DR CI.
For different values of the sample size and the underlying performance gap, we compute the rejection rate of the statistical test across 1,000 runs.
As before, the classifier $\sfA$ represents the oracle classifier that has the optimal decision boundary, which is linear, but the classifier $\sfB$ now uses a linear decision boundary that is shifted from the optimal one by a fixed amount, thereby shifting $\Delta^{\sfA\sfB}$ away from zero.
As such, $\sfB$ performs increasingly worse as $\Delta^{\sfA\sfB}$ increases.

\begin{figure}[t]
    \centering
    \includegraphics[height=0.3\textwidth]{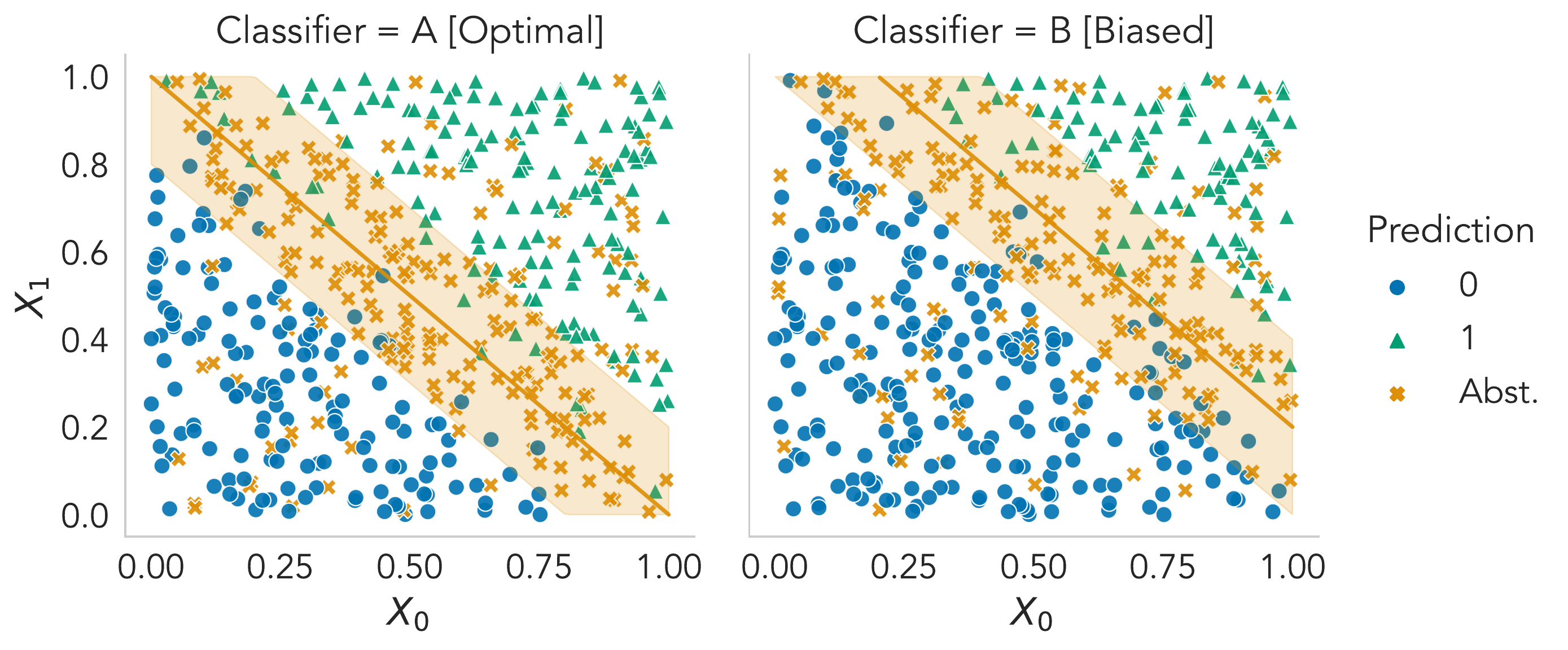} 
    \caption{A simulated example for the power experiment in which $\Delta^{\sfA\sfB} = 0.123$. 
    The evaluation data is the same as the one in Figure~\ref{fig:binary_mar}.
    For $\sfB$, the decision boundary of $\sfA$ is shifted diagonally upwards by $\mu=0.2$; in the power experiment, we experiment with various values of $\mu$ (and thus $\Delta^{\sfA\sfB}$).
    }
    \label{fig:binary_mar_power}
\end{figure}

To increasingly vary the counterfactual score difference between two classifiers, we set $\sfA$ as the same classifier as in Section~\ref{sec:binary_mar_setup} and set $\sfB$ to use the (optimal) linear decision boundary of $\sfA$ shifted diagonally by a fixed amount $\mu$.
Specifically, $f^\sfB(x_0, x_1) = \sigma(x_0+x_1 - (1+\mu))$.
An example with $\mu=0.2$ is shown in Figure~\ref{fig:binary_mar_power}.
While $\Delta^{\sfA\sfB}$ is not strictly a linear function of $\mu$, it is gradually increasing as $\mu$ increases, as shown in Table~\ref{tbl:mu_delta}.
Aside from this difference, both classifiers use the same abstention mechanism as classifier $\sfA$ from the previous experiment, and the data generating process is also identical to the previous experiment.

\begin{figure}[tb]
    \centering
    \includegraphics[width=0.5\textwidth]{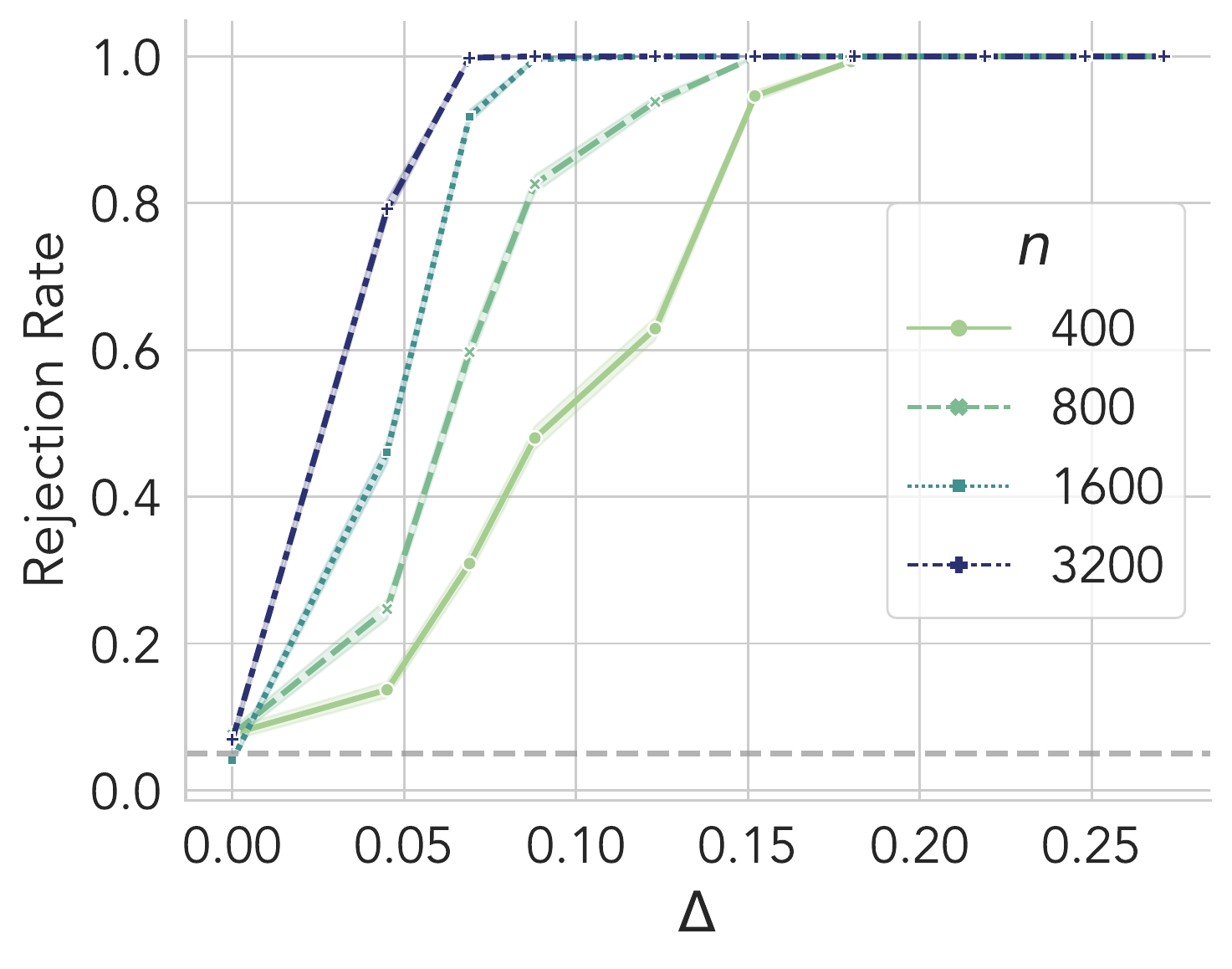}
    \caption{Power of the statistical test for $H_0: \Delta^{\sfA\sfB} = 0$ derived by our 95\% DR CIs, plotted for different values of $n$ (sample size) and $\Delta^{\sfA\sfB}$, which varies based on the distance between the (linear) decision boundaries of $\sfA$ and $\sfB$.
    Mean rejection rates of $H_0$ over 1,000 simulations are shown, with 1 standard error as shaded error bars.
    As either $n$ or $\Delta^{\sfA\sfB}$ grows large, the power approaches 1.}
    \label{fig:power} 
\end{figure}

Figure~\ref{fig:power} plots the rejection rates of the level-$\alpha$ statistical test, for $\alpha=0.05$, against different values of $\Delta^{\sfA\sfB}$ (0 to 0.27) for various sample sizes ($n=400, 800, 1600, 3200$).
Here, we plot the miscoverage rate as a function of the resulting values of $\Delta^{\sfA\sfB}$ directly. 
We use the super learner to learn the nuisance functions.
Overall, we see that as $n$ or $\Delta^{\sfA\sfB}$ increases, the power of the statistical test quickly approaches 1, implying that the test can consistently detect a gap in counterfactual scores if either the sample size or the difference gets large.

On a 128-core CPU machine, using parallel processing, the entire compute time it took to produce Figure~\ref{fig:power} was approximately 88 minutes.

\iftoggle{compact}{
\begin{table}[t]
    \centering
    \caption{The relationship between $\Delta^{\sfA\sfB}$ and $\mu$, the distance between the linear decision boundaries of $\sfA$ and $\sfB$, in the power experiment of Section~\ref{sec:power}.}
    \vspace{\baselineskip}
    \begin{tabular}{c|ccccccccccc}
        \toprule
        $\Delta^{\sfA\sfB}$ & 0.0 & 0.045 & 0.069 & 0.088 & 0.123 & 0.152 & 0.180 & 0.181 & 0.219 & 0.248 & 0.271 \\  \midrule
        $\mu$               &  0 & 0.05 & 0.10 & 0.15 & 0.20 & 0.25 & 0.30 & 0.35 & 0.40 & 0.45 & 0.50 \\
        \bottomrule
    \end{tabular}
    \label{tbl:mu_delta}
\end{table}
}{
\begin{table}[t]
        \centering
        \caption{The relationship between $\Delta^{\sfA\sfB}$ and $\mu$, the distance between the linear decision boundaries of $\sfA$ and $\sfB$, in the power experiment of Section~\ref{sec:power}.}
        \label{tbl:mu_delta}
        \begin{tabular}{c|ccccccccccc}
            \toprule
            $\Delta^{\sfA\sfB}$ & 0.0 & 0.045 & 0.069 & 0.088 & 0.123 & 0.152 & 0.180 & 0.181 & 0.219 & 0.248 & 0.271 \\  \midrule
            $\mu$               &  0 & 0.05 & 0.10 & 0.15 & 0.20 & 0.25 & 0.30 & 0.35 & 0.40 & 0.45 & 0.50 \\
            \bottomrule
        \end{tabular}
\end{table}

}

\iftoggle{compact}{\clearpage}{}

\subsection{Details on the CIFAR-100 experiment}\label{sec:cifar100_details}

The abstaining classifiers compared in the experiments are variants of the VGG-16 CNN model with batch normalization~\citep{simonyan2015very}.
Specifically, the feature representation layers are obtained from a model\footnote{\url{https://github.com/chenyaofo/pytorch-cifar-models}} trained on the training set of the CIFAR-100 dataset and are fixed during evaluation.
Using half ($n=5,000$) of the validation set, we train a L2-regularized softmax output layer and its softmax response (SR) for the abstention mechanism.
The comparison is done on the other half ($n=5,000$) of the validation set.
This version of the VGG-16 features and the softmax layer is used for all scenarios, with different abstention mechanisms described in the main text, except for the last comparison, where we compare this softmax layer with VGG-16's original 3-layer output model (2 hidden layers of size 512).

The nuisance functions, $\hat\pi$ and $\hat\mu_0$ for each classifier in each scenario, also utilize the pre-trained representations of the VGG-16 layer, but their output layers (both L2-regularized linear models) are trained separately via cross-fitting.

The pre-trained VGG-16 features on the CIFAR-100 validation set were first obtained using a single NVIDIA A100 GPU, taking approximately 20 seconds. 
On a 128-core CPU machine, using parallel processing, the rest of the computation to produce Table~\ref{tbl:cifar100} took less than 10 seconds (note that there are no repeated runs in this experiment).

\subsection{Sensitivity to different positivity levels}\label{sec:positivity_experiment}

\begin{figure}[t]
    \centering
    \iftoggle{compact}{
        \includegraphics[width=0.4\textwidth]{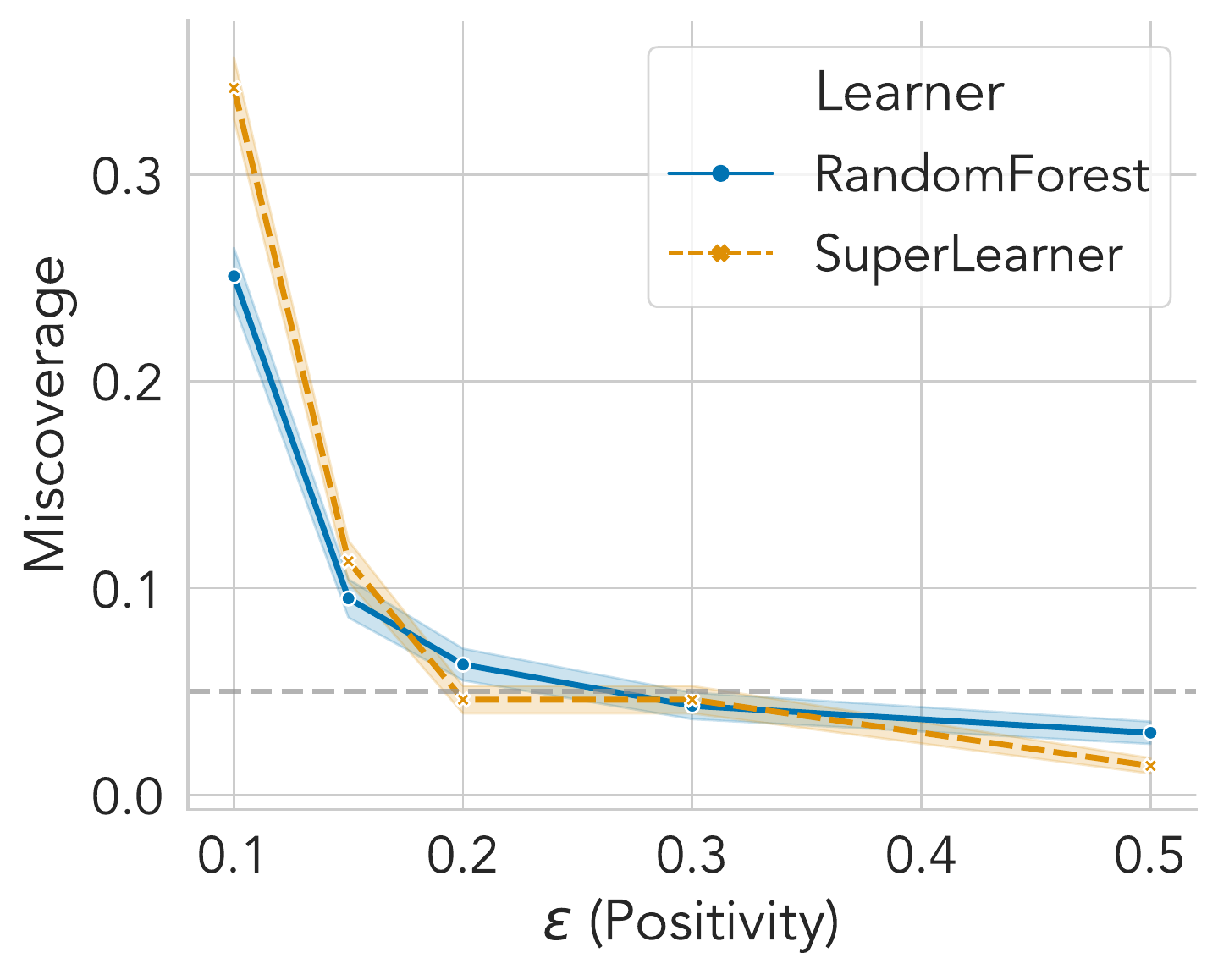} 
    }{
        \includegraphics[width=0.5\textwidth]{contents/figures/simulated/binary_mar_eps0.2/miscoverage_by_positivity.pdf}
    }
    \caption{Miscoverage rates of 95\% doubly robust CIs by varying the level of $\epsilon$ (positivity), plotted for different nuisance function learners. Each point is the mean over 1,000 repeated simulations; shaded error bars represent 1 standard error.}
    \label{fig:positivity}
\end{figure}

Here, we examine how the DR estimator is affected by the level of positivity, i.e., $\epsilon$ in~\eqref{assn:positivity_cfscore}.
As discussed in the main paper, positivity violations make it infeasible to properly identify and estimate causal estimands.
In practice, we expect the DR estimator to remain valid up until $\epsilon$ becomes smaller than a certain (small) number. 
To validate this, we use the same setting from our first experiment (Section~\ref{sec:simulated}; Appendix~\ref{sec:binary_mar_setup}) but vary the level of positivity from $\epsilon=0.5$ (MCAR) to $\epsilon=0.1$ (positivity near-violation).

Figure~\ref{fig:positivity} plots the miscoverage rate of the DR estimator, averaged over 1,000 repeated simulations, using the three nuisance learner choices we used in Section~\ref{sec:simulated}.
The result confirms that the DR estimator, when using either the random forest or the super learner, retains validity as long as $\epsilon \geq 0.2$, in this particular case; as $\epsilon$ shrinks to below $0.2$, the miscoverage rates start to go above the significance level. 
This confirms that there is a (problem-dependent) level of positivity we must expect for the DR estimator to work; otherwise, we do not expect the counterfactual target to be a meaningfully identifiable quantity in the first place.

On a 128-core CPU machine, using parallel processing, the entire compute time it took to produce Figure~\ref{fig:positivity} was approximately 12 minutes.

\end{document}